\documentclass{tlp}
\nocite{*}

\usepackage{latexsym}
\usepackage{verbatim}
\usepackage{url}

\newcommand\bcmdtab{\noindent\bgroup\tabcolsep=0pt%
  \begin{tabular}{@{}p{10pc}@{}p{20pc}@{}}}
\newcommand\ecmdtab{\end{tabular}\egroup}

\newcommand{\nnot}{\textit{not }}

\long\def\COMMENT#1\ENDCOMMENT{\message{(Commented text...)}\par}

\def\st{\vspace{1mm}\noindent}

\def\ni{\noindent}
\def\and{ \ \wedge}

\def\iff{\;\Leftrightarrow\;}

\def\beq{\begin{equation}}
\def\eeq#1{\label{#1}\end{equation}}

\def\initially{\; \hbox{\bf initially} \;}

\def\causes{\; \hbox{\bf causes} \;}

\def\executable{\; \hbox{\bf executable\_if} \;}

\def\always{\; \hbox{\bf always } \; }
\def\iif{\; \hbox{\bf if} \; }

\def\calb{{\cal B}}

\def\calf{{\cal F}}

\def\calpp{{\cal PP}}

\def\calS0{{\cal S}_0}

\def\la{{\leftarrow}}

\def\lan{\langle}
\def\ran{\rangle}

\def\until{\hbox{\bf until}}
\def\always{\hbox{\bf always}}
\def\eventually{\hbox{\bf eventually}}
\def\next{\hbox{\bf next}}
\def\goal{\hbox{\bf goal}}

\def\naf{{not} \;}
\def\theequation{\arabic{equation}}

\newcommand{\belist}{\begin{list}{$\bullet$}{\topsep=1pt \parsep=0pt \itemsep=1pt}}
\newcommand{\enlist}{\end{list}}

\newtheorem{example}{Example}
\newtheorem{theorem}{Theorem}
\newtheorem{proposition}{Proposition}
\newtheorem{lemma}{Lemma}

\newtheorem{definition}{Definition}

\title
[Planning with Preferences using Logic Programming]
{Planning with Preferences using Logic Programming\footnote{
This paper is an extended version of a paper
that appeared in the proceedings of the 7$^{th}$ International
Conference on Logic Programming and Non-Monotonic Reasoning, 2004.
}}

\author[Tran Cao Son  and  Enrico Pontelli]
{
    TRAN CAO SON and ENRICO PONTELLI \\
    Knowledge Representation, Logic, and  Advanced Programming Laboratory \\
    Computer Science Department,
         New Mexico State University, Las Cruces, New Mexico, USA\\
         \email{tson@cs.nmsu.edu} and 
         \email{epontell@cs.nmsu.edu} 
}

\submitted{29 June 2004}
\revised{17 December 2004, 9 August 2005}
\accepted{29 August 2005}

\begin{document}
\maketitle

\begin{abstract}
We present a declarative language,  ${\cal PP}$,
for the high-level specification of preferences between possible solutions 
(or trajectories) of a planning problem. This novel language 
allows users to elegantly express non-trivial, multi-dimensional 
preferences and priorities over such preferences. The 
semantics of ${\cal PP}$ 
allows the identification of \emph{most preferred trajectories}
for a given goal. 
We also provide an answer set programming
implementation of planning problems with ${\cal PP}$ preferences. 
\end{abstract}


\section{Introduction and Motivation}

Planning---in its classical sense---is the problem of finding 
a sequence of actions that achieves a predefined goal \cite{Rei01}. 
Most of the research
in AI planning has been focused on methodologies and issues related 
to the development of efficient planners. To date, several 
efficient planning 
systems have been developed---e.g.,
see \cite{aips02}.
These developments can be attributed to 
the discovery of good domain-independent heuristics, the use of 
domain-specific knowledge, and the development of efficient data 
structures used in the implementation of  planning algorithms. 
Logic programming has played a significant role in this line 
of research, providing a declarative framework for the encoding
of different forms of knowledge and its effective use during the
planning process \cite{sbm02a}.

However, relatively  limited effort has been placed on 
addressing  several 
important aspects in real-world planning domains, such as 
{\em plan quality} and {\em preferences about plans}. In many real
world frameworks, the space of feasible plans to achieve the
goal is dense,
but many of such plans, even if executable,  may present 
undesirable features. In these frameworks, it may be simple
to find a solution (\emph{``a''} plan); rather, the challenge, is
to produce a solution that is considered satisfactory w.r.t.
the \emph{needs} and \emph{preferences} of the user.
Thus, feasible plans may have a measure
of quality, and only a subset of them may be considered acceptable.
These issues can be seen in the following example.

\begin{example}
\label{exp1}
Let us consider planning problems in the {\em travel domain}.
A planning problem  in this domain 
can be represented by the following elements\footnote{
   Precise formulae will be presented later.
}:

\begin{itemize}
\item a set of fluents of the form 
$at(l)$, where $l$ denotes a location, 
such as {\em home, school, neighbor, airport, etc.}; 

\item an initial location $l_i$; 

\item a final location $l_f$; and

\item a set of actions of the form $method(l_1,l_2)$
where $l_1$ and $l_2$ are two distinct locations and 
$method$ is one of the available transportation methods, such 
as {\em drive, walk, ride\_train, bus, taxi, fly, bike, etc.}
In addition, there might be conditions that restrict the 
applicability of actions in certain situations. For example,
one can ride a taxi only if the taxi has been called, which can
be done only if one has some money; one can fly from one place 
to another if he/she has the ticket; etc. 
\end{itemize}

Problems in this domain are often rich in solutions
because of the large number of actions which can be used 
in the construction of a plan. Consider, for example, 
a simple situation, in which a user wants to construct a 3-leg 
trip, that starts from a location $l_1$ and ends at $l_4$, 
and there are 10 ways to move along each leg, one of 
them being the action $walk(l_i,l_{i+1})$. The number 
of possible plans is $10^3$ and 
\[
walk(l_1,l_2), walk(l_2,l_3), walk(l_3, l_4)
\]
is a possible plan that achieves the goal.
In most of the cases, the user is likely to 
dismiss this plan and selects another one for various reasons;
among them the total distance from $l_1$ to $l_4$ might be too 
large, the time and/or energy required to complete the plan would 
be too much, etc.
This plan, however, would be a reasonable one, and most likely the only 
acceptable solution, for someone wishing to visit his/her neighbors.

In selecting the plan  deemed appropriate for him/herself,
the user's preferences play an important role. For example,
a car-sick person would prefer walking over driving whenever the action 
{\em walk} can be used. A wealthy millionaire cannot afford to waste too 
much time and would prefer to use a taxi. A poor student would 
prefer to bike over riding a taxi, simply because he cannot afford the
taxi. 
Yet, the car-sick person will have to ride a taxi whenever 
other transportations are not available;
the millionaire will have to walk whenever no taxi is available;
and the student will have to use a taxi when he does not have time. 
In other words, there are instances where a user's preference might not 
be satisfied and he/she will have to use plans that do not
satisfy such preference. 
\hfill$\Box$
\end{example}

The above discussion shows that users' preferences play a
decisive role 
in the choice of a plan. It also shows that hard-coding  user
preferences as a part of the goal is not a satisfactory way to 
deal with preferences. Thus,
we need to be able to evaluate plan components
at a finer  granularity than simply as consistent or violated.
In \cite{myers99a}, it is argued that users' preferences 
are of vital importance in selecting a plan for execution
when the planning problem has too many solutions. 
It is worth observing that, with a few exceptions---like
the system SIPE-2 with meta-theoretic biases 
\cite{myers99a}---most planning systems do not allow 
users to specify their preferences and use them in finding 
plans. The responsibility in selecting the most appropriate plan 
rests solely on the users. It is also important to
observe that \emph{preferences} are different from  \emph{goals}
in  a planning problem: a plan \emph{must} satisfy the goal, while
it may or may not satisfy the preferences. The
distinction is analogous to the separation between
\emph{hard} and \emph{soft} constraints \cite{bista1}.
For example, let us consider a user with the {\em goal} 
of being at the airport who {\em prefers} to use a taxi 
over driving his own car; considering his preference as 
a soft constraint, then the user will have to drive his car 
to the airport if no taxi is available; on the other hand,
if the preference is considered as a hard constraint, no plan will
achieves the user's goal when no taxi is available.

In this paper,
we will investigate the problem of integrating user preferences
into a planner. We will develop a \emph{high-level
 language} for the
specification of user preferences, 
and then provide a logic programming encoding of
the language, based
on Answer Set Programming \cite{smodels-constraint}. 
As demonstrated in this work,
normal logic programs with answer set semantics \cite{gel90a} 
provide a natural and elegant framework to effectively handle
planning with preferences. 

We divide the preferences that a user might have
in different categories:
\belist
\item \emph{Preferences about a state:} the user prefers to be in 
a state $s$ that satisfies a property $\phi$ rather than 
a state $s'$ that does not satisfy it, in case
both lead to the satisfaction of his/her goal;
\item \emph{Preferences about an action:} the user prefers
to perform 
the action $a$, whenever it is feasible and it
allows the goal to be achieved;
\item \emph{Preferences about a trajectory:} the user prefers a 
trajectory that satisfies a certain property $\psi$ 
over those that do not satisfy this property; 
\item \emph{Multi-dimensional Preferences:} the user has a 
\emph{set} of 
preferences, with an ordering
among them. A trajectory satisfying a more 
favorable preference is given priority over 
those that satisfy less favorable preferences. 
\enlist
It is important to observe
 the difference between $\phi$ and $\psi$ in the 
above definitions. $\phi$ is a \emph{state} property,
whereas $\psi$ is a formula over the whole \emph{trajectory}
(from the initial state to the state that satisfies the
given goal).

\medskip
The rest of this paper is organized as follows. 
In Section \ref{sec2}, we review the foundations
of answer set planning. Section \ref{sec3}
presents the high-level preference 
language $\cal PP$. Section \ref{sec4} 
describes a methodology to compute preferred trajectories using answer 
set planning. In Section \ref{related} we discuss the related  work,
while Section \ref{sec5} presents the final discussion and conclusions.

\section{Preliminary -- Answer Set Planning}
\label{sec2}

In this section we review the basics of planning using 
logic programming with answer set 
semantics---\emph{Answer Set Planning (or ASP)} 
\cite{dimo97,lif02a,sub95}. We will assume that the 
effect of actions on the world and the relationship between
fluents in the world are expressed in an appropriate 
language. In this paper, we make use of the ontologies 
of a \emph{variation} of the action description 
language $\calb$ \cite{gl98}. 
In this language, an action theory is defined over two 
disjoint sets of names---the set of 
{\em actions} {\bf A} and the set of {\em fluents} {\bf F}.
An action theory  is a pair $(D,I)$, where 
\begin{itemize}
	\item $D$ is a set of propositions expressing 
		the effects of actions,
		the relationship between fluents, and
		the executability conditions for the actions\footnote{
	Executability conditions were not originally included in 
	the definition of the language $\calb$ in \cite{gl98}. 
}; 
	\item $I$ is a set of propositions representing 
		the initial state of the world.
\end{itemize}
Instead of presenting a formal definition of $\calb$, we 
introduce the syntax of the language by presenting an action 
theory representing the travel domain of Example \ref{exp1}. 
We write 
\begin{itemize}
\item $at(l)$, where $l$ is 
a constant representing a possible location, such as 
{\em home, airport, school, neighbor, bus\_station}, to 
denote the fact that the agent\footnote{Throughout the paper, 
we assume that we are working in 
   a single agent (or user) environment. 
   Fluents and actions with variables are shorthand representing 
   the set of their ground instantiations. 
} is at the location $l$; 
\item $available\_car$ to denote the fact that the car 
is available for the agent's use;
\item $has\_ticket(l_1,l_2)$ to denote the fact 
that the agent has the ticket 
to fly from $l_1$ to $l_2$; etc.
\end{itemize}
The action of driving from location $l_1$ to 
location $l_2$ causes the agent 
to be at the location $l_2$ and is represented in $\calb$ by the 
following {\em dynamic causal law}:
\[
drive(l_1,l_2) \causes at(l_2) \iif at(l_1).
\]
This action can only be executed if the car is available 
for the agent's use at the location $l_1$ and
there is a road connecting $l_1$ and $l_2$. 
This information is represented by an {\em
executability condition}:
\[
drive(l_1,l_2) \executable available\_car, at(l_1), road(l_1,l_2). 
\]
The fact that one can only be at one location at a time is 
represented by the following {\em static causal law} ($l_1\neq l_2$):
\[
\neg at(l_2) \iif at(l_1).
\]
Other actions with their executable conditions and effects 
are represented in a similar way.

To specify the fact that the agent is initially at home, he has 
some money, and
a car is available for him to use, we write 
\[
\begin{array}{c}
\initially(at(home)) \\
\initially(has\_money) \\
\initially(available\_car(home)) \\
\end{array}
\]

\begin{example}
\label{exxp}
Below, we list some more actions with their effects and 
executability conditions, using $\calb$.
\[
\begin{array}{lllll}
walk(l_1,l_2) & \causes & at(l_2) & \iif & at(l_1), road(l_1,l_2) \\ 
bus(l_1,l_2) & \causes & at(l_2) & \iif & at(l_1),  road(l_1,l_2)  \\
flight(l_1,l_2) & \causes & at(l_2) & \iif & at(l_1), has\_ticket(l_1,l_2)\\
take\_taxi(l_1,l_2) & \causes & at(l_2) & \iif & at(l_1), road(l_1,l_2) \\
buy\_ticket(l_1,l_2) & \causes &  has\_ticket(l_1,l_2) \\
call\_taxi(l) & \causes & available\_taxi(l) & \iif & has\_money\\
rent\_car(l) & \causes & available\_car(l) & \iif & has\_mony \\ 
bus(l_1,l_2) & \executable & has\_money \\
flight(l_1,l_2) & \executable & connected(l_1,l_2) \\
take\_taxi(l_1,l_2) & \executable & available\_taxi(l_1)\\
buy\_ticket(l_1,l_2)& \executable & has\_money \\ 
\end{array}
\]
where the $l$'s denote locations, airports, or 
bus stations.
The fluents and actions are self-explanatory.\hfill$\Box$
\end{example}

Since our main concern in this paper is not the language 
for representing actions and their effects, we omit here the 
detailed definition of the proposed variation of
 $\calb$ \cite{gl98}. It suffices for us 
to remind the readers that the semantics of an action theory 
is given by the notion of \emph{state} and by a 
\emph{transition function} $\Phi$, that specifies the 
result of the execution of an action $a$ in a state $s$
(denoted by $\Phi(a,s)$). Each state $s$ is a set of fluent literals 
satisfying the two properties: 
\begin{enumerate}
\item  for every fluent $f \in \mathbf{F}$, either 
$f\in s$ or $\neg f \in s$ 
but $\{f, \neg f\} \not\subseteq s$; and 
\item  $s$ satisfies the static causal laws. 
\end{enumerate}
A state $s$ satisfies a fluent literal $f$ ($f$ holds in $s$), denoted 
by $s \models f$, if $f \in s$.  A state $s$ satisfies a static causal law 
\[
f \iif p_1,\ldots,p_n
\]
if, whenever $s \models p_i$ for every $1 \le i \le n$, then
we have that $s \models f$. An action $a$ is {\em executable} in a state 
$s$ if there exists an executability condition
\[
a \executable p_1,\ldots,p_n
\] 
in $D$ such that $s \models p_i$ for every $i$, $\le i \le n$. 
An action theory $(D,I)$ is \emph{consistent} if 
\begin{enumerate}
\item $s_0 = \{f \mid \initially(f) \in I\}$ is a state, 
and 
\item for every action $a$ and state $s$
such that $a$ is executable in $s$, we have that $\Phi(a,s) \ne \emptyset$. 
\end{enumerate}
In this paper, we will assume that $(D,I)$ is consistent. 
A \emph{trajectory} of an action theory $(D,I)$ is a 
sequence
$s_0a_1s_1\ldots a_{n}s_n$ where $s_i$'s are states,
$a_i$'s are actions, and
$s_{i+1} \in \Phi(s_i, a_{i+1})$ for $i \in \{0,\ldots,n-1\}$.

A planning problem is specified by a triple $\langle D,I,G \rangle$,
where $(D,I)$ is an action theory and $G$ is a fluent formula 
(a propositional formula constructed from fluent literals and
propositional connectives)
representing the goal. A possible solution to $\langle D,I,G \rangle$
is a trajectory $\alpha = s_0 a_1 s_1 \ldots a_m s_m$,
where  $s_0 \models I$ and $s_m \models G$.
 In this case, we say that the trajectory $\alpha$
achieves $G$.

\medskip
Answer set planning \cite{dimo97,lif02a,sub95} solves a planning problem 
$\lan D, I, G \ran$ by translating it into a logic program
$\Pi(D, I, G)$ which consists of
\emph{(i)} rules describing $D$, $I$, and $G$; and
\emph{(ii)} rules generating action occurrences.
It also has a parameter, {\em length}, 
declaring the maximal length of the trajectory
that the user can accept.
The two key predicates of $\Pi(D, I, G)$ are:
\begin{itemize}
\item $holds(f,t)$ -- the
	fluent literal $f$ holds at the time moment $t$; and
\item $occ(a,t)$ -- the action $a$ occurs at the time moment $t$.
\end{itemize}
$holds(f,t)$ can be extended to define $holds(\phi,t)$ 
for an arbitrary fluent formula $\phi$, which states that $\phi$
holds at the time moment $t$. Details about the program 
$\Pi(D, I, G)$ can be found in \cite{sbm02a}\footnote{A 
Prolog program for translation $\langle D, I, G\rangle$ 
into $\Pi(D, I, G)$ can be found at 
\url{http://www.cs.nmsu.edu/~tson/ASPlan/Preferences/translate.pl}.
}. 
The key property of the translation of
$\langle D, I, G \rangle$ into $\Pi(D, I, G)$  is that it ensures 
that each trajectory achieving $G$ corresponds to an answer
set  of $\Pi(D, I, G)$, and each answer set
of $\Pi(D, I, G)$ corresponds to a trajectory achieving $G$.

\begin{theorem} \cite{sbm02a}
\label{th1}
For a planning problem $\lan D,I,G \ran$ with a consistent
action theory $(D,I)$ and  maximal plan length $n$, 
\begin{enumerate}
\item
if $s_0a_1\ldots a_{n}s_n$ is a trajectory achieving $G$, then
there exists an answer set $M$ of $\Pi(D,I,G)$ such that:
	\begin{enumerate}
	\item $occ(a_i,i-1) \in M$ for  $i \in \{1,\ldots,n\}$, and
	\item $s_i = \{f \mid holds(f,i) \in M\}$ for $i \in \{0,\ldots,n\}$.
	\end{enumerate}
\item
if $M$ is an answer set of $\Pi(D,I,G)$, then there exists an integer
$0 \le k \le n$ such that $s_0 a_{1}\ldots a_{k}s_k$ is a
trajectory achieving $G$, where $occ(a_{i},i-1) \in M$ for $1
\le i \le k$ and $s_i = \{f \mid holds(f,i) \in M\}$ 
for $i \in \{0,\ldots,k\}$.
\end{enumerate}
\end{theorem}
In the rest of this work, if $M$ is an answer set of
$\Pi(D,I,G)$, then we will denote with $\alpha_M$ the
trajectory achieving $G$ represented by $M$.
Answer sets of the program $\Pi(D,I,G)$ can be computed using 
answer set solvers such as {\bf smodels} \cite{smodels-ai}, 
{\bf dlv} \cite{eiter98a}, {\bf cmodels} \cite{cmodels2},
{\bf ASSAT} \cite{assat},
and {\bf jsmodels} \cite{jsmodels}. 

\section{A Language for Planning Preferences Specification}
\label{sec3}

In this section, we introduce the language $\calpp$
for planning preferences specification. This language 
allows users to express their preferences among plans
that achieve the same goal. 
We subdivide preferences in different classes: 
\emph{basic desires}, \emph{atomic preferences}, and 
\emph{general preferences}. Intuitively, a basic desire 
is a preference expressing a desirable property of 
a plan such as the use of certain action over the others, 
the satisfaction of a fluent formula,
or a temporal property (Subsection \ref{sec31}). 
An atomic preference describes a one-dimensional ordering on plans
and allows us to describe a ranking over the plans given a set of possibly  
conflicting preferences (Subsection \ref{sec32}). 
Finally, a general preference provides means for users to 
combine different preference dimensions
(Subsection \ref{sec33}).

Let $\langle D,I,G \rangle$ be a planning problem 
with the set of actions {\bf A} and the set of fluents {\bf F}; 
let $\calf_F$ be the set of all fluent formulae over {\bf F}.
The language $\calpp$ is defined as special formulae  over {\bf A} and 
{\bf F}. We will illustrate the
different types of preferences using the action theory 
representing the travel domain discussed earlier
(Example \ref{exxp}). User preferences about plans in this domain 
are often based on properties of actions. 
Some of these properties are flying is {\em 
very fast} but {\em very expensive}; walking is {\em slow}, 
and {\em very tiring} if the distance between the two locations is large 
but {\em cheap}; driving is {\em tiring} and {\em costs a little}
but it is {\em cheaper} than flying and {\em faster} than walking;
etc. 

\subsection{Basic Desires}
\label{sec31}
 
A basic desire is a formula expressing a single preference about a 
trajectory. Consider a user who is at {\em home} and 
wants to go to {\em school} (goal) spending as little money as possible
(preference),
i.e., his desire is to save money. He has only three alternatives:
{\em walking, driving}, or {\em take\_taxi}. Walking is the 
cheapest and riding a taxi is the most expensive. Thus, 
a preferred trajectory for him should contain 
the action {\em walk(.,.)}. This preference 
could also be expressed by a formula that forbids the fluent 
{\em available\_taxi(home)} or {\em available\_car} to become 
true in every state of the trajectory, thus preventing him 
to drive or take a taxi to school. 
These two alternatives of preference representation are not 
always equivalent. The first one represents the desire of
leaving a state using a specific group of actions, while 
the second one represents the desire of being in certain 
states. 

Basic desires are constructed by using {\em state desires}
and/or {\em goal preferences}. 
Intuitively, a state desire describes a basic user
preference to be considered in the context of a specific
state. A state desire $\varphi$ (where $\varphi$ is 
a fluent formula) implies
that we prefer a state $s$ such that $s \models \varphi$. A state
desire $occ(a)$ implies that we prefer to leave 
state $s$ using the action $a$. In many cases, 
it is also desirable to talk about
the final state of the trajectory---we call this a {\em goal 
preference}.
These cases are formally  defined next.
\begin{definition}[State Desires and Goal Preferences]
\label{def1}\label{def2}
A (primitive) {\em state desire} is
either a formula $\varphi$, where $\varphi \in \calf_F$, or
a formula of the form $occ(a)$, where $a \in \mathbf{A}$.

\noindent
A \emph{goal preference} is a formula of the form 
$\textbf{goal}(\varphi)$, where $\varphi$ is a formula 
in $\calf_F$.
\end{definition}

We are now ready to define a basic desire that expresses a user
preference over the trajectory. As such, in addition to the 
propositional connectives $\wedge, \vee, \neg$, 
we will also use the temporal connectives 
$\next$, $\always$, $\until$, and $\eventually$. 
\begin{definition}[Basic Desire Formula]
\label{def3}
A \emph{basic desire formula} is a formula satisfying
one of the following conditions:
\begin{list}{$\bullet$}{\topsep=2pt \itemsep=1pt \parsep=0pt}
\item  a goal preference $\varphi$ is a basic desire formula;
\item a state desire $\varphi$ is a basic desire formula;
\item given the basic desire formulae $\varphi_1, \varphi_2$, then
$\varphi_1 \wedge \varphi_2$, $\varphi_1 \vee \varphi_2$,
$\neg \varphi_1$, 
$\next(\varphi_1)$, $\until(\varphi_1,\varphi_2)$, 
$\always(\varphi_1)$, and 
$\eventually(\varphi)$ are also basic desire formulae.
\end{list}
\end{definition}
For example, to express the fact that a user would like to take the taxi or 
the bus to go to school, we can write:
\[
\eventually(\:occ(bus(home,school)) \vee occ(taxi(home,school))\:).
\]
If the user's desire is not to call a taxi,  we can write
\[
\always(\:\neg occ(call\_taxi(home))\:).
\]
If for some reasons, the user's desire is not to see any taxi around his 
home, the desire 
\[
\always(\:\neg available\_taxi(home)\:).
\]
can be used. Note that these encodings have different consequences---the
second  prevents taxis to be present independently from whether
it was called or not.

The definition above is used to develop formulae expressing 
a desire regarding the structure of trajectories. In the next definition,
we will describe when a trajectory satisfies a basic desire formula.
In a later section (Section \ref{implementation}), 
we will present logic programming rules that can be 
added to the program $\Pi(D,I,G)$ to compute trajectories that 
satisfy a basic desire. In 
the following definitions, given a trajectory
$\alpha = s_0 a_1 s_1 \cdots a_n s_n$, the notation $\alpha[i]$ denotes
the trajectory $s_i a_{i+1} s_{i+1} \cdots a_n s_n$.

\begin{definition}[Basic Desire Satisfaction]
\label{def4}
\label{basicsat}
Let $\alpha = s_0 a_1 s_1 a_2 s_2 \cdots a_n s_n$ be a trajectory, and
let $\varphi$ be a basic desire formula. $\alpha$ satisfies $\varphi$
(written as $\alpha \models \varphi$) iff
one of the following holds
\begin{list}{$\bullet$}{\topsep=2pt \itemsep=1pt \parsep=0pt}
\item $\varphi = \textbf{goal}(\psi)$ and $s_n \models \psi$
\item $\varphi = \psi \in \calf_F$ and $s_0 \models \psi$
\item $\varphi = occ(a)$\/, $a_1 = a$\/, and $n \geq 1$
\item $\varphi = \psi_1 \wedge \psi_2$\/, $\alpha \models \psi_1$ and
		$\alpha \models \psi_2$
\item $\varphi = \psi_1 \vee \psi_2$\/, $\alpha \models \psi_1$ or
	$\alpha \models \psi_2$
\item $\varphi = \neg \psi$ and $\alpha \not\models \psi$
\item $\varphi = \next(\psi)$\/,  $\alpha[1] \models \psi$\/, and $n\geq 1$
\item $\varphi = \always(\psi)$ and $\forall (0 \leq i \leq n)$
		we have that $\alpha[i] \models \psi$
\item $\varphi = \eventually(\psi)$ and $\exists (0 \leq i \leq n)$ such
		that $\alpha[i] \models \psi$
\item $\varphi = \until(\psi_1,\psi_2)$ and 
$\exists (0 \leq i \leq n)$ such 
that $\alpha[j] \models \psi_1$ for all 
$0 \leq j < i$ and  $\alpha[i] \models \psi_2$.
\end{list}
\end{definition}
Definition \ref{basicsat} allows us to check whether a trajectory 
satisfies a basic desire. This will also allow us to compare 
trajectories.
Let us start with the simplest form of trajectory preference, involving
a single desire. 

\begin{definition}[Ordering Between Trajectories w.r.t. Single Desire]
\label{def5}
Let $\varphi$ be a basic desire formula and let $\alpha$ and
$\beta$ be two trajectories. The trajectory $\alpha$ is
{\em preferred} to the trajectory $\beta$ (denoted as
$\alpha \prec_{\varphi} \beta$) if 
$\alpha \models \varphi$ and $\beta \not\models \varphi$.

We say that
$\alpha$ and $\beta$ are {\em indistinguishable w.r.t. $\varphi$}
(denoted as $\alpha \approx_{\varphi} \beta$) if one of the
two following cases occur: 
\begin{enumerate}
\item  $\alpha \models \varphi$ and $\beta \models \varphi$, or
\item 
$\alpha \not\models \varphi$ and $\beta \not\models \varphi$.
\end{enumerate}
\end{definition}
Whenever it is clear from the context, we will omit
$\varphi$ from $\prec_{\varphi}$ and $\approx_{\varphi}$.
We will also allow a weak form of single preference, described next.
\begin{definition}[Weak Single Desire Preference]
\label{def7}
Let $\varphi$ be a basic desire formula and let $\alpha, \beta$
be two trajectories. $\alpha$ is
{\em weakly preferred} to $\beta$ (denoted $\alpha \preceq_{\varphi} \beta$)
iff  $\alpha \prec_{\varphi} \beta$ or $\alpha \approx_{\varphi} \beta$.
\end{definition}
It is easy to see that $\approx_\varphi$ is an equivalence relation over 
the set of trajectories.

\begin{proposition}
Given a basic desire $\varphi$, the relation $\approx_{\varphi}$ is an
equivalence relation.
\end{proposition}
\begin{proof}
\begin{enumerate}
\item \emph{Reflexivity:} this case is obvious.
\item \emph{Symmetry:} let us assume that, given two trajectories
	$\alpha, \beta$ we have that $\alpha \approx_{\varphi} \beta$. This
	implies that either both trajectories satisfy $\varphi$ or neither
	of them do. Obviously, if $\alpha \models \varphi$ and $\beta \models
	\varphi$ ($\alpha \not\models \varphi$ and $\beta \not\models \varphi$) then
	we have also that $\beta \models \varphi$ and $\alpha \models
	\varphi$ ($\beta \not\models \varphi$ and $\alpha \not\models \varphi$), which
	leads to $\beta \approx_{\varphi} \alpha$.
\item \emph{Transitivity:} let us assume that for the trajectories $\alpha, \beta, \gamma$
	we have that 
	\[ \alpha \approx_{\varphi} \beta \hspace{1cm}\textnormal{ and } \hspace{1cm} \beta \approx_{\varphi} \gamma \]
	From the first component, we have two possible cases:
	\begin{enumerate}
	\item $\alpha \models\varphi$ and $\beta\models \varphi$. Since $\beta \approx_{\varphi} \gamma$,
	we need to have $\gamma \models \varphi$, which leads to $\alpha \approx_{\varphi} \gamma$.
	\item $\alpha \not\models \varphi$ and $\beta \not\models \varphi$. This second component, together
		with $\beta \approx_{\varphi} \gamma$ leads to $\gamma \not\models \varphi$, and thus 
		$\alpha \approx_{\varphi} \gamma$.
	\end{enumerate}
\end{enumerate}
\end{proof}
 
In the next proposition, we will show that 
$\preceq_\varphi$ is a partial order over the set of 
equivalence classes representatives of $\approx_\varphi$\footnote{
   This means that $\preceq_\varphi$ satisfies the following three 
   properties: 
(i) Reflexivity: $\alpha \preceq_\varphi \alpha$;
(ii) Antisymmetry: if
$\alpha \preceq_\varphi \beta$ and 
$\beta \preceq_\varphi \alpha$ then $\alpha \approx_\varphi \beta$; and
(iii) Transitivity: if
$\alpha \preceq_\varphi \beta$ and 
$\beta \preceq_\varphi \gamma$ then $\alpha \preceq_\varphi \gamma$
where $\alpha$, $\beta$, and $\gamma$ are arbitrary trajectories. 
}. 

\begin{proposition}
\label{p1}
The relation $\preceq_{\varphi}$ defines a partial order over the
set of representatives of the equivalence classes of $\approx_{\varphi}$.
\end{proposition}
\begin{proof}
Let us prove the three properties. 

\begin{enumerate}
\item \emph{Reflexivity:} consider a representative $\alpha$. 
	Since either $\alpha \models \varphi$ or $\alpha \not\models \varphi$,
	we have that $\alpha \preceq_{\varphi} \alpha$.
\item \emph{Anti-symmetry:} consider two representatives
	$\alpha, \beta$ and let us assume that 
	$\alpha \preceq_{\varphi} \beta$ and $\beta \preceq_{\varphi} \alpha$.
	Since both $\alpha$ and $\beta$ are equivalent class 
	representatives of $\approx_\varphi$, to prove this property, 
	it suffices to show that $\alpha \approx_{\varphi} \beta$. 
	First of all, we can observe that from $\alpha \preceq_{\varphi} \beta$
	we have either $\alpha \prec_{\varphi} \beta$ 
	or $\alpha \approx_{\varphi} \beta$. If
	$\alpha \prec_{\varphi} \beta$ then this means that 
	$\alpha \models \varphi$ and $\beta \not\models \varphi$. But 
	this would imply that 
	$\beta \not\preceq_{\varphi} \alpha$. Then
	we must have that $\alpha \approx_{\varphi} \beta$. 

\item \emph{Transitivity:} consider three representatives
	$\alpha_1, \alpha_2, \alpha_3$ and let us assume
\[ \alpha_1 \preceq_{\varphi} \alpha_2 \wedge \alpha_2 \preceq_{\varphi} \alpha_3 \]
	Let us consider two cases.

\begin{itemize}
\item $\alpha_1 \prec_{\varphi} \alpha_2$. This implies 
	that $\alpha_1 \models \varphi$ and 
	$\alpha_2 \not\models \varphi$. 
	Because $\alpha_2 \preceq_{\varphi} \alpha_3$, we have that 
	$\alpha_3 \not\models \varphi$. This, together with 
	$\alpha_1 \models \varphi$, allows us
	to conclude $\alpha_1 \prec_{\varphi} \alpha_3$. 

\item $\alpha_1 \approx_{\varphi} \alpha_2$, 
	then either
	$\alpha_1 \models \varphi$ and $\alpha_2 \models \varphi$, or
	$\alpha_1 \not\models \varphi$ and $\alpha_2 \not\models \varphi$.
	In the first case, we have $\alpha_1 \approx_{\varphi} \alpha_3$ 
	if $\alpha_3 \models \varphi$ and 
	$\alpha_1 \prec_{\varphi} \alpha_3$ 
	if $\alpha_3 \not\models \varphi$, i.e., 
	$\alpha_1 \preceq_{\varphi} \alpha_3$. 
	If instead we have 
	the second possibility, then since $\alpha_2 \not\models \varphi$ 
	and  $\alpha_2 \preceq_{\varphi} \alpha_3$, 
	we must have $\alpha_3 \not\models \varphi$. 
	This allows us to conclude that 
	$\alpha_1 \approx_{\varphi} \alpha_3$ and 
	thus $\alpha_1 \preceq_{\varphi} \alpha_3$.
\end{itemize}

\end{enumerate}
\end{proof}
We next define the notion of most preferred trajectories.

\begin{definition}
[Most Preferred Trajectory w.r.t. Single Desire]
\label{def8}
Let $\varphi$ be a basic desire formula. A trajectory $\alpha$ is 
said to be a {\em most preferred trajectory} w.r.t. $\varphi$ 
if there is no trajectory $\beta$ such that 
$\beta \prec_\varphi \alpha$.
\end{definition}
Note that in the presence of preferences, 
a most preferred trajectory might 
require extra actions that would have been otherwise 
considered unnecessary. 
\begin{example}
\label{ex3}
Let us enrich the action theory
of Example \ref{exxp} with an action called 
{\em buy\_coffee}, which allows one to have coffee, i.e, 
the fluent {\em has\_coffee}  becomes true.
The coffee is not free, i.e., the agent will have to pay
some money if he buys coffee. 
This action can only be executed at the nearby Starbucks shop. 
If our agent wants to be at school and 
prefers to have coffee, we write:
\[
\goal(\textnormal{\em has\_coffee}).
\]
Any plan satisfying this preference requires
the agent to stop at the Starbucks shop before 
going to school. 
E.g., while $s_0 walk(home,school) s_1$,
where $s_0$ and $s_1$ denote the initial state (the agent is at home)
and the final state (the agent is at school), respectively, 
is a valid trajectory 
for the agent to achieve his goal, 
this is not a most preferred trajectory; instead,
the agent has to go to the Starbucks shop, buy the coffee,
and then go to school. Besides the action of {\em buy\_coffee} 
that is needed for him to get the coffee, the most preferred
trajectory requires the action of {\em going to the 
coffee shop}, which is not necessary if he does not 
have the preference of having the coffee. 

Observe that the most preferred trajectory contains the 
action {\em buy\_coffee}, which can only be executed when 
the agent has some money. As such, if the agent does not have 
any money, this action will not be executable and no 
trajectory achieving the goal of being at the school will 
contain this action. This means that no plan 
can satisfy the agent's preference, i.e., he will have 
to go to school without coffee.
\hfill$\Box$
\end{example}

The above definitions are also expressive enough to describe
a significant portion of preferences that frequently occur in real-world 
domains. Since some of them are particularly important, we will introduce 
some syntactic sugar to simplify their use:

\begin{list}{$\bullet$}{\topsep=2pt \itemsep=2pt \parsep=2pt}
\item (Strong Desire)
	given the basic desire formulae $\varphi_1, \varphi_2$, 
	 $\:\:\varphi_1 < \varphi_2$ denotes
	$\varphi_1 \wedge \neg \varphi_2$. 

\item (Weak Desire)
	given the basic desire formulae $\varphi_1, \varphi_2$, 
	$\:\:\varphi_1 <^w \varphi_2$ denotes
	$\varphi_1 \vee \neg \varphi_2$.
\item (Enabled Desire) given two actions $a_1, a_2$, we will denote
	with $a_1 <^e a_2$ the formula 
	$(executable(a_1) \wedge executable(a_2)) \Rightarrow 
		(occ(a_1) < occ(a_2))$
	where $$
	executable(a) = \bigvee_{a \executable p_1,\ldots,p_k} p_1 \wedge \ldots \wedge p_k.$$
\end{list}
In the rest of the paper, we often use the following shorthands:
\begin{list}{$\bullet$}{\topsep=2pt \itemsep=2pt \parsep=2pt}
\item For a sequence of preference formulae $\varphi_1,\ldots,\varphi_k$, 
	$$\varphi_1 < \ldots < \varphi_k$$ stands 
for 
\[\bigwedge_{i \in \{1,\ldots,k-1\}} (\varphi_i < \varphi_{i+1}).\]

\item For a sequence of preference formulae $\varphi_1,\ldots,\varphi_k$, 
	$$\varphi_1 <^w \ldots <^w \varphi_k$$ stands 
for 
\[
\bigwedge_{i \in \{1,\ldots,k-1\}} (\varphi_i <^w \varphi_{i+1}).
\]

\item For the sequence of actions $a_1,\ldots,a_k,b_1,\ldots,b_m$,
	$$
	(a_1 \vee \ldots \vee a_k) <^e (b_1 \vee \ldots \vee b_m)
	$$ is a shorthand for 
	$$
	\bigwedge_{i \in \{1,\ldots,k\}, \: j \in \{1,\ldots,m\}} (a_i <^e b_j).
	$$

\item
For actions with parameters like {\em drive} or {\em walk}, we sometime write 
$drive <^e walk$ to denote the preference 
\[
\bigvee_{l_1, l_2 \in S, \: l_1 \ne l_2} (drive(l_1, l_2) <^e walk(l_1, l_2)).
\]
where $S$ is a set of pre-defined locations. Intuitively, this preference states 
that we prefer to drive rather than to walk between locations belonging to the
set $S$. For example, if $S = \{home, school\}$ then this preference says 
that we prefer to drive from home to school and vice versa. 
\end{list}
We can prove the following simple property of $\leq^w$.
\begin{lemma} \label{l1}
Consider the set of basic desire formulae and let us interpret
$<^w$ as a relation. This relation is transitive.
\end{lemma}
\begin{proof}
Let $\varphi_1 <^w \varphi_2$ and
$\varphi_2 <^w \varphi_3$. But these are the same as
\[(\varphi_1 \vee \neg \varphi_2) \wedge (\varphi_2 \vee \neg \varphi_3) \]
which implies
\[ \varphi_1 \vee \neg \varphi_3\]
and thus $\varphi_1 <^w \varphi_3$.
\end{proof}

\subsection{Atomic Preferences}
\label{sec32}
Basic desires allow the users to specify their preferences and 
can be used in selecting trajectories which satisfy them. From 
the definition of a basic desire formula, we can assume that users 
always have a set of desire formulae and that their intention is to 
find a trajectory that satisfies all such formulae. In many 
cases, this proves to be too strong and results 
in  situations where no preferred trajectory can be found.
Consider again the preference in Example \ref{ex3}, it is 
obvious that the user cannot have a plan that costs him 
nothing and yet satisfies his preferences. 
In the travel domain, {\em time} and {\em cost}
are two criteria that a user might have when making a 
travel plan. These two criteria are often in conflict, 
i.e., a transportation method that takes little time often costs 
more. As such, it is very unlikely that the user can get a plan 
that costs very little and takes very little time. 

\begin{example}
\label{ex4}
Let us continue with our travel domain. Again, let us 
assume that the agent is at home and he wants to go to school.
To simplify the representation, we will write $bus$,
$taxi$, $drive$, and $walk$ to say that the agent takes
the bus, taxi, drive, or walk to school, respectively. The following 
is a desire expressing that the agent
prefers to get the fastest possible way to go to school
(assume that both {\em driving} and {\em taking the bus}
require about the same amount of time to reach the school):
\[
time = \always(taxi <^e (drive \vee bus) <^e walk)
\]
On the other hand, when the agent is not in a hurry,
he/she prefers to get the cheaper way to go to school
(assume that {\em driving} and {\em taking the bus}
cost about the same amount of money):
\[
cost = \always(walk <^e (drive \vee bus) <^e taxi)  
\]
Here, the preference states that the agent 
prefers to execute first the action that consumes the least amount 
of money.
\hfill$\Box$
\end{example}
It is easy to see that any trajectories satisfying the preference
{\em time} will not satisfy the preference {\em cost} and vice 
versa. This discussion shows that it is necessary to provide 
users with a simple way to \emph{rank} their basic desires. 
To address the problem, we allow a new type of formulae,
called {\em atomic preferences}, which represents 
an ordering between basic desire formulae.
\begin{definition}[Atomic Preference]
\label{def9}
An \emph{atomic preference formula} is defined as a formula
of the type
$ \varphi_1 \lhd \varphi_2 \lhd \cdots \lhd \varphi_n $ 
where $\varphi_1, \dots, \varphi_n$ are basic desire formulae.
\end{definition}
The intuition behind an atomic preference is to provide an 
ordering between different desires---i.e., it indicates that
trajectories that satisfy 
$\varphi_1$ are preferable
to those that satisfy $\varphi_2$, etc. Clearly, basic
desire formulae are special cases of atomic preferences---where
all preference formulae have the same rank.
The definitions of $\approx$ and $\prec$ can now be extended
to compare trajectories w.r.t. atomic preferences. 
\begin{definition}[Ordering Between Trajectories 
w.r.t. Atomic Preferences]
\label{def10}
Let $\alpha, \beta$ be two trajectories,  and let 
$\Psi = \varphi_1 \lhd \varphi_2 \lhd \cdots \lhd \varphi_n$
be an atomic preference formula. 
\belist
\item $\alpha, \beta$ are {\em indistinguishable} w.r.t. $\Psi$ 
(written as
$\alpha \approx_{\Psi} \beta$) if 
$$\forall i. \: ( \: 1 {\leq} i {\leq} n \Rightarrow  
\alpha \approx_{\varphi_i} \beta\: ) \:.$$  
\item $\alpha$ is {\em preferred} to $\beta$ w.r.t. $\Psi$ 
(written as $\alpha \prec_{\Psi} \beta$)
if $\exists (1 \leq i \leq n)$ such that
\begin{enumerate}
\item $\forall (1 \leq j < i)$ we have that $\alpha \approx_{\varphi_j} \beta$,
and 
\item  $\alpha \prec_{\varphi_i} \beta$.
\end{enumerate}
\enlist
We will say that $\alpha \preceq_{\Psi} \beta$ if either
$\alpha \prec_{\Psi} \beta$ or $\alpha \approx_{\Psi} \beta$.
\end{definition}
It is easy to see that $\approx_{\Psi}$ is an equivalence relation 
on the set of trajectories. The following proposition 
is similar to Proposition \ref{p1}. 

\begin{proposition}
\label{p2}
For an atomic preference $\Psi$, $\preceq_{\Psi}$ is
a partial order over the set of representatives of the
 equivalence classes of $\approx_{\Psi}$.
\end{proposition}

\begin{proof}
Let us analyze the three properties.

\begin{itemize}
\item \emph{Reflexivity:} Consider a representative $\alpha$. 
	By Definition \ref{def10}, $\alpha \approx_{\Psi} \alpha$, which
	leads to $\alpha \preceq_{\Psi} \alpha$.

\item \emph{Anti-symmetry:} Let $\alpha \preceq_{\Psi} \beta$ and
	$\beta \preceq_{\Psi} \alpha$. 
	Again, it is enough if we can show that 
	$\alpha \approx_{\Psi} \beta$. 
	Let us assume, by contradiction,
	that $\alpha \prec_{\Psi} \beta$. This means that there is
	a value of $i$ such that, for all $1 \leq j < i$ we have
	that $\alpha\approx_{\varphi_j} \beta$ and 
	$\alpha \prec_{\varphi_i} \beta$. But this implies that 
	$\beta \approx_{\varphi_j} \alpha$ for $j < i$ and
	$\beta \not\prec_{\varphi_i} \alpha$, which ultimately
	means $\beta \not\preceq_{\Psi} \alpha$, contradicting the
	initial assumptions. 

\item \emph{Transitivity:} let $\alpha_1, \alpha_2, \alpha_3$ be
	three representatives such that
\[ \alpha_1 \preceq_{\Psi} \alpha_2 \wedge \alpha_2 \preceq_{\Psi} \alpha_3 \]
	Let us consider the possible cases arising from the first component:
	\begin{itemize}
	\item  $\alpha_1 \approx_{\Psi} \alpha_2$. This means 
			that $\alpha_1\approx_{\varphi_j} \alpha_2$ 
			for all $1\leq j \leq n$. We have two sub-cases:
		\begin{itemize}
		\item $\alpha_2 \approx_{\Psi} \alpha_3$. Because 
		$\approx_\Psi$ is an equivalence relation, we have that 
		$\alpha_1 \approx_{\Psi} \alpha_3$, which implies that
		$\alpha_1 \preceq_{\Psi} \alpha_3$.
		\item $\alpha_2 \prec_{\Psi} \alpha_3$. This means that 
		there exists $i$, $1 \le i \le n$, 
		such that $\alpha_2 \approx_{\varphi_k} \alpha_3$ 
		for all $1\leq k < i$ and
		$\alpha_2 \prec_{\varphi_i} \alpha_3$. Since 
		$\approx_{\varphi_j}$ is an equivalence 
		relation, we have that 
		$\alpha_1 \approx_{\varphi_j} \alpha_3$ for all  
		$1 \le j < i$. Furthermore, 
		$\alpha_1 \approx_{\varphi_i} \alpha_2$
		and $\alpha_2 \prec_{\varphi_i} \alpha_3$ imply
		that $\alpha_1 \models \varphi_i$,
		$\alpha_2 \models \varphi_i,$ and 
		$\alpha_3 \not\models \varphi_i$. Thus, 
		$\alpha_1 \prec_{\varphi_i} \alpha_3$. Hence, 
		$\alpha_1 \preceq_{\Psi} \alpha_3$.
		\end{itemize}

	\item $\alpha_1 \prec_{\Psi} \alpha_2$. This implies that there 
			exists $i$, $1 \le i \le n$, 
		such that $\alpha_1 \approx_{\varphi_k} \alpha_2$ 
		for all $1\leq k < i$ and
		$\alpha_1 \prec_{\varphi_i} \alpha_2$. Again, we have two sub-cases: 
		\begin{itemize}
		\item $\alpha_2 \approx_{\Psi} \alpha_3$. This means 
		that $\alpha_2 \approx_{\varphi_j} \alpha_3$ for all $j$,
		$1 \le j \le n$. So, we have that 
		$\alpha_1 \approx_{\varphi_j} \alpha_3$ for all  
		$1 \le j < i$, since $\approx_{\varphi_j}$ is an equivalence 
		relation. Similar to the above case, we can show that 
		$\alpha_1 \prec_{\varphi_i} \alpha_2$ and 
		$\alpha_2 \approx_{\varphi_i} \alpha_3$ 
		implies that 
		$\alpha_1 \prec_{\varphi_i} \alpha_3$. Thus,
		$\alpha_1 \prec_{\Psi} \alpha_3$.
		\item $\alpha_2 \prec_{\Psi} \alpha_3$. This means that 
		there exists $j$, $1 \le j \le n$, 
		such that $\alpha_2 \approx_{\varphi_k} \alpha_3$ 
		for all $1\leq k < j$ and
		$\alpha_2 \prec_{\varphi_j} \alpha_3$. 
		If $j \ge i$, we have that $\alpha_1 \approx_{\varphi_k} \alpha_3$		
		for $k < i$ and $\alpha_1 \prec_{\varphi_i} \alpha_3$	
		(because 
			$\alpha_1 \prec_{\varphi_i} \alpha_2$
		and     $\alpha_2 \approx_{\varphi_i} \alpha_3$).
		Otherwise, if $j < i$, using this fact and 
		the transitivity of $\approx_{\varphi_k}$, we can conclude 
		that  $\alpha_1 \approx_{\varphi_k} \alpha_3$ 
		for all $1\leq k < j$ and
		$\alpha_1 \prec_{\varphi_j} \alpha_3$, which implies that 
		$\alpha_1 \prec_{\Psi} \alpha_3$.
		\end{itemize}
	\end{itemize}
\end{itemize}
\end{proof}

\begin{definition}[Most Preferred Trajectory w.r.t. Atomic Preferences]
A trajectory $\alpha$ 
is {\em most preferred} if there is no 
other trajectory that is preferred to $\alpha$.
\end{definition}

\begin{example}
\label{ex41}
Let us continue the Example \ref{ex4}. 
The two preferences $time$ and $cost$ can be combined into different atomic 
preferences, e.g., 
\[
time \lhd cost \;\;\;\;\;\; \textnormal{ or } \;\;\;\;\;\;
cost \lhd time.
\]
The first one is more appropriate for the agent when he is in a hurry 
while the second one is more appropriate for him when he has time. The 
trajectory \[ \alpha = s_0\: walk(home,school)\: s_1\] 
is preferred to the trajectory 
\[\beta = s_0\: call\_taxi(home)\: s_1'\: taxi(home, school)\: s_2'\] 
w.r.t. to the preference $cost \lhd time$,
i.e., $\alpha \prec_{cost \lhd time} \beta$.\footnote{For brevity, 
we omit the description of the states $s_i$'s.}
On the other hand, we have that $\beta \prec_{time \lhd cost} \alpha$.
\hfill$\Box$
\end{example}

\subsection{General Preferences}
\label{sec33}

Atomic preferences allow users to list their preferences 
according to their importance, where more preferred desires appear 
before less preferred ones. Naturally, when a user 
has a set of atomic preferences, there is a need for combining them 
to create a new preference that can be used to select the best possible 
trajectory suitable to him/her. This can be seen in the next example. 

\begin{example}
\label{exp5}
Let us continue with the action theory described in Example \ref{exp1}.
Besides {\em time} and {\em cost}, agents often have their preferences
based on the level of comfort and/or safety of the the available 
transportation methods. This preferences can be represented by the 
formulae 
\[
comfort = \always(flight <^e (drive \vee bus) <^e walk)
\]
and
\[
safety  = \always(walk <^e flight <^e (drive \vee bus)).
\]
Now, consider an agent who has in mind the four basic desires 
{\em time, cost, comfort,} and {\em safety}. He can rank these 
preferences and create different atomic preferences, i.e.,
different orders among these preferences. Let us assume that 
he has combined these four desires and produced the following
two atomic preferences 
\[
\Psi_1 = comfort \lhd safety \:\:\: \textnormal{ and } \:\:\:
\Psi_2 = cost \lhd time.
\]
Intuitively, $\Psi_1$ is a comparison between level of comfort and 
safety, while $\Psi_2$ is a comparison between affordability
and duration.

Suppose that the agent would like to 
combine $\Psi_1$ and $\Psi_2$ to create 
a preference stating that he prefers trajectories that
are as comfortable as possible and cost as little as possible.
So far, the only possible way for him to combine these 
two preferences is to concatenate them in a certain 
order and view the result as a new atomic preference. However,
neither $\Psi_1 \lhd \Psi_2$ nor $\Psi_2 \lhd \Psi_1$ meets
the desired criteria---as they simply state that $\Psi_1$ is
more relevant than $\Psi_2$ (or vice versa). The only way to 
accomplish the desired effect is to decompose $\Psi_1$ and
$\Psi_2$ and rebuild a more complex atomic preference.
This shows that the agent might have to define a new atomic
preference for his newly introduced preference.
\hfill$\Box$
\end{example}
The above discussion shows that it is necessary to provide additional ways 
for combining atomic preferences. This is the topic of this sub-section. 
We will introduce general preferences, which 
can be used to describe a multi-dimensional order among preferences. 
Formally, we define general preferences as follows.
\begin{definition}[General Preferences]
A \emph{general preference formula} is a formula satisfying one of the
following conditions:
\begin{list}{$\bullet$}{\topsep=2pt \itemsep=1pt \parsep=0pt}
\item An atomic preference $\Psi$ is a general preference;
\item If $\Psi_1, \Psi_2$ are general preferences, then
		$\Psi_1 \& \Psi_2$, $\Psi_1 \mid \Psi_2$,
                 and 
		$!\: \Psi_1$ are general preferences;
\item If $\Psi_1, \Psi_2, \dots, \Psi_k$ is a 
        collection of general preferences, then
	$\Psi_1 \lhd \Psi_2 \lhd \cdots \lhd \Psi_k$ is
	 a general preference.
\end{list}
\end{definition}
In the above definition, the operators $\&, \mid, !$ are used to express
different ways to combine preferences. Syntactically, 
they are similar to the operations $\wedge, \vee, \neg$ employed
in  the construction of basic desire formulae. Semantically,
they differ from the operations $\wedge, \vee, \neg$ 
in a subtle way. Intuitively, the constituents of a general preference 
are atomic preferences; and a general preference provides a 
means for combining different orderings among trajectories created 
by its constituents, thus providing a means for the selection 
of a most preferred trajectory. Let us consider the case where 
a general preference has two constituents $\Psi_1$ and $\Psi_2$.
As we will see later, each preference will induce two relations on 
trajectories, the indistinguishable and preferred relations, 
as in the case of atomic preferences. 
In other words, $\Psi_i$ can be represented by this pair 
of relations. 
Given a general preference $\Psi_i$, 
let $o_i$ and $e_i$ denote the set of pairs of trajectories 
$(\alpha, \beta)$ such that $\alpha \prec_{\Psi_i} \beta$
and $\alpha \approx_{\Psi_i} \beta$, respectively. 
The operators
$\&, \mid, !$ look at this characterization and define two 
relations among trajectories that satisfy the following equations:
\begin{itemize}
\item For the formula $\Psi_1 \& \Psi_2$, we define 
	$$(o_1,e_1) \& (o_2,e_2) \stackrel{\textnormal{\footnotesize def}}{=} 
        ((o_1 \cap o_2), (e_1 \cap e_2)). 
        $$
	This says that the relation representing the ordering between trajectories 
	induced by $\&$ is created as the intersection of the relations 
	representing the orderings between trajectories induced by its 
	components. In this case, 
	a most preferred trajectory is the one which is most preferred
	w.r.t. every component of the formula.

\item For the formula $\Psi_1 \mid \Psi_2$, we define 
	$$(o_1,e_1) \mid (o_2,e_2) \stackrel{\textnormal{\footnotesize def}}{=} 
        ((o_1 \cap o_2) \cup (o_1 \cap e_2) \cup (e_1 \cap o_2), (e_1 \cap e_2)). 
       $$
	Here, the relation induced by $\mid$ guarantees that the most 
	preferred trajectory is the one which is most preferred 
	w.r.t. at least one component of the formula and it is indistinguishable 
	from others w.r.t. the remaining component of the formula.

\item For the formula $!\Psi_1$, we define 
	$$!(o_1,e_1) \stackrel{\textnormal{\footnotesize def}}{=} 
        (({o_1}^{-1} \cup e_1), e_1) 
       $$
	which basically reverses the relations induced by $\Psi_1$.
\end{itemize}
This is made more precise in the following definition. 
\begin{definition}[Ordering Between Trajectories w.r.t. General Preferences]
\label{def11}
Let $\Psi$ be a general preference and let $\alpha, \beta$ be two
trajectories. We say that 
\begin{list}{$\bullet$}{\topsep=2pt \itemsep=1pt \parsep=0pt}
\item 
$\alpha$ is {\em preferred} to $\beta$, denoted by $\alpha \prec_{\Psi} \beta$,
 if:
\begin{enumerate}
\item $\Psi$ is an atomic preference and $\alpha \prec_{\Psi} \beta$
\item $\Psi = \Psi_1 \& \Psi_2$ and
	$\alpha \prec_{\Psi_1} \beta$ and 
	$\alpha \prec_{\Psi_2} \beta$
\item $\Psi = \Psi_1 \mid \Psi_2$ and: 
	\begin{enumerate}
	\item
		$\alpha \prec_{\Psi_1} \beta$ and 
		$\alpha \approx_{\Psi_2} \beta$; or
	\item  $\alpha \prec_{\Psi_2} \beta$ and 
		$\alpha \approx_{\Psi_1} \beta$; or
	\item $\alpha \prec_{\Psi_1} \beta$ and 
		$\alpha \prec_{\Psi_2} \beta$
	\end{enumerate}

\item $\Psi = \:! \Psi_1$ and 
        $\beta \prec_{\Psi_1} \alpha$

\item $\Psi = \Psi_1 \lhd \cdots \lhd \Psi_k$, and there
	exists $1 \leq i \leq k$ such that:
	\emph{(i)} $\forall (1 \leq j < i)$ we have that $\alpha \approx_{\Psi_j} \beta$, and \emph{(ii)} $\alpha \prec_{\Psi_i} \beta$.
\end{enumerate}
\item
$\alpha$ is {\em indistinguishable} from $\beta$,
denoted by $\alpha \approx_{\Psi} \beta$, if:
\begin{enumerate}
\item $\Psi$ is an atomic preference and $\alpha \approx_{\Psi} \beta$.
\item $\Psi = \Psi_1 \& \Psi_2$, $\alpha \approx_{\Psi_1} \beta$,
	$\alpha \approx_{\Psi_2} \beta$.
\item $\Psi = \Psi_1 \mid \Psi_2$, $\alpha \approx_{\Psi_1} \beta$, and
	$\alpha \approx_{\Psi_2} \beta$.
\item $\Psi = \:! \Psi_1$ and $\alpha \approx_{\Psi_1} \beta$.
\item $\Psi = \Psi_1 \lhd \cdots \lhd \Psi_k$,
	and for all $1 \leq i \leq k$ we have that 
	$\alpha \approx_{\Psi_i} \beta$. 
\end{enumerate}
\end{list}
\end{definition}
Similar as above, $\alpha \preceq_{\Psi} \beta$ indicates that
either $\alpha \prec_{\Psi} \beta$ or $\alpha \approx_{\Psi} \beta$.
Before we move on, let us observe that
the formula $\Psi_1 \lhd \dots \lhd \Psi_n$, 
where each $\Psi_i$ is a basic desire, can be viewed as both 
an atomic preference as well as a general preference. This is not 
ambiguous since the semantic definition for the two cases 
coincide. 
It is easy to see that the following holds for $\preceq_{\Psi}$.

\begin{lemma} \label{l2}
For every pair of trajectories $\alpha$ and $\beta$ and 
a general preference $\Psi$ such that $\alpha \preceq_\Psi \beta$, 

\begin{list}{$-$}{\topsep=1pt \parsep=0pt \itemsep=1pt}
\item if $\Psi = \Psi_1 \& \Psi_2$ then $\alpha \preceq_{\Psi_1} \beta$ and
	$\alpha \preceq_{\Psi_2} \beta$.
\item if $\Psi = \Psi_1 \mid \Psi_2$ then $\alpha \preceq_{\Psi_1} \beta$ and
	$\alpha \preceq_{\Psi_2} \beta$.
\item if $\Psi = \:! \Psi_1$ then $\beta \preceq_{\Psi_1} \alpha$.
\end{list}
\end{lemma}
\noindent
We can also show that for every general preference $\Psi$,
$\approx_{\Psi}$ is an equivalence relation over trajectories 
and $\preceq_{\Psi}$ is a partial order on the 
set of representatives of the equivalence classes  of $\approx_{\Psi}$.
To prove this property, we need the following lemma.

\begin{lemma} \label{sideq}
Let $\Psi$ be a general preference formula and let $\alpha$, $\beta$, and $\gamma$
be three trajectories. It holds that 
\begin{itemize}
\item 
if $\alpha \prec_{\Psi} \beta$
and $\beta \approx_{\Psi} \gamma$ then $\alpha \prec_{\Psi} \gamma$; and 
\item 
if $\alpha \prec_{\Psi} \beta$
and $\alpha \approx_{\Psi} \gamma$ then $\gamma \prec_{\Psi} \beta$.
\end{itemize}
\end{lemma}

\begin{proof}
Let us prove the result by structural induction on $\Psi$. Because 
the proof of the two items are almost the same, we present below the 
proof of the first item.

\begin{itemize}

\item \emph{Base:} If $\Psi$ is an atomic preference, 
and $\Psi= \varphi_1 \lhd \cdots \lhd \varphi_k$ then from 
$\alpha \prec_{\Psi} \beta$ (Definition \ref{def10}) 
we obtain that there exists $1 \leq i \leq k$ such that 
	\begin{itemize}
	\item $\alpha \approx_{\varphi_j} \beta$ for all $1 \leq j < i$ and
	\item $\alpha \prec_{\varphi_i} \beta$. 
	\end{itemize}
Furthermore, from $\beta \approx_{\Psi} \gamma$ we have that
	$\beta \approx_{\varphi_j} \gamma$ for all $1 \leq j \leq k$.
Since $\approx_{\varphi_j}$ are all equivalence relations, we 
obtain that $\alpha \approx_{\varphi_j} \gamma$ for all $1 \leq j < i$; 
furthermore, since $\alpha \prec_{\varphi_i} \beta$, then
$\alpha\models \varphi_i$ and $\beta \not\models \varphi_i$. Since 
$\beta \approx_{\varphi_i} \gamma$ then necessarily we have also that
$\gamma \not\models \varphi_i$. This allows us to conclude that 
	$\alpha \prec_{\Psi} \gamma$.

\item \emph{Inductive Step:}

\begin{enumerate}
\item $\Psi = \Psi_1 \& \Psi_2$. Since
$\alpha \prec_{\Psi} \beta$, we have 
	$\alpha \prec_{\Psi_1} \beta$ and 
	$\alpha \prec_{\Psi_2} \beta$. Furthermore, 
	$\beta \approx_{\Psi} \gamma$ implies
	$\beta \approx_{\Psi_1} \gamma$ and 
	$\beta \approx_{\Psi_2} \gamma$. From the induction
	hypothesis we have
	$\alpha \prec_{\Psi_1} \gamma$ and 
	$\alpha \prec_{\Psi_2} \gamma$, which leads to
	$\alpha \prec_{\Psi} \gamma$.

\item $\Psi = \Psi_1 \mid \Psi_2$. From 
	$\alpha \prec_{\Psi} \beta$ we have three possible
	cases:
	\begin{enumerate}
	\item $\alpha \prec_{\Psi_1} \beta$ and 
		$\alpha \prec_{\Psi_2} \beta$. In this case,
		since $\beta \approx_{\Psi} \gamma$ implies
		$\beta \approx_{\Psi_1} \gamma$ and 
		$\beta \approx_{\Psi_2} \gamma$, we 
		have that $\alpha \prec_{\Psi_1} \gamma$ and
		$\alpha \prec_{\Psi_2} \gamma$. This implies that
		$\alpha \prec_{\Psi} \gamma$.

	\item $\alpha \prec_{\Psi_1} \beta$ and 
		$\alpha \approx_{\Psi_2} \beta$. From $\beta \approx_{\Psi}
		\gamma$ we obtain $\beta \approx_{\Psi_1} \gamma$ and
		$\beta \approx_{\Psi_2} \gamma$. Since $\approx_{\Psi_2}$
		is an equivalence relation, we can infer
		$\alpha \approx_{\Psi_2} \gamma$. Furthermore, from the
		induction hypothesis we obtain $\alpha \prec_{\Psi_1} \gamma$.
		These two conclusions lead to $\alpha \prec_{\Psi}
		\gamma$.

	\item $\alpha \prec_{\Psi_2} \beta$ and 
		$\alpha \approx_{\Psi_1} \beta$. This case is 
			symmetrical to the previous one.
	\end{enumerate}

\item $\Psi = ! \Psi_1$. From 
$\alpha \prec_{\Psi} \beta$ we obtain $\beta \prec_{\Psi_1} \alpha$.
Since $\beta \approx_{\Psi} \gamma$ implies 
	$\beta \approx_{\Psi_1} \gamma$, from the induction hypothesis
	we can conclude $\gamma \prec_{\Psi_1} \alpha$ and ultimately
	$\alpha \prec_{\Psi} \gamma$.

\item $\Psi = \Psi_1 \lhd \cdots \lhd \Psi_k$. From the 
definition of $\alpha \prec_{\Psi} \beta$ we have that
there exists $1 \leq i \leq k$ such that
\begin{enumerate}
\item $\alpha \approx_{\Psi_j} \beta$ for all $1 \leq j < i$ and
\item $\alpha \prec_{\Psi_i} \beta$
\end{enumerate}
Furthermore, from $\beta \approx_{\Psi} \gamma$ we know that
	$\beta \approx_{\Psi_j} \gamma$ for all $1 \leq j \leq k$.
Since $\approx_{\Psi_j}$ are all equivalence relations, we have
that $\alpha \approx_{\Psi_j} \gamma$ for all $1 \leq j < i$. 
Furthermore, from the induction hypothesis, from 
$\alpha \prec_{\Psi_i} \beta$ and $\beta \approx_{\Psi_i} \gamma$ we
can conclude $\alpha \prec_{\Psi_i} \gamma$. This finally leads to
$\alpha \prec_{\Psi} \gamma$.
\end{enumerate}
\end{itemize}
\end{proof}

\begin{lemma} \label{l4}
Let $\Psi$ be a general preference formula and $\alpha$, $\beta$, and $\gamma$
be three trajectories. It holds that 
if $\alpha \prec_{\Psi} \beta$
and $\beta \prec_{\Psi} \gamma$ then $\alpha \prec_{\Psi} \gamma$.
\end{lemma}

\begin{proof}
Let us prove the result by structural induction on $\Psi$. 

\begin{itemize}

\item \emph{Base:} If $\Psi$ is an atomic preference, 
and $\Psi= \varphi_1 \lhd \cdots \lhd \varphi_k$ then from 
$\alpha \prec_{\Psi} \beta$ (Definition \ref{def10}) 
we obtain that there exists $1 \leq i \leq k$ such that 
	\begin{itemize}
	\item $\alpha \approx_{\varphi_j} \beta$ for all $1 \leq j < i$ and
	\item $\alpha \prec_{\varphi_i} \beta$. 
	\end{itemize}
Furthermore, from $\beta \prec_{\Psi} \gamma$, we know that 
there exists $1 \leq l \le k$ such that 
	\begin{itemize}
	\item $\beta \approx_{\varphi_j} \gamma$ for all $1 \leq j < l$ and
	\item $\beta \prec_{\varphi_l} \gamma$. 
	\end{itemize}
If $i \le l$, it is easy to see that $\alpha \approx_{\varphi_j} \gamma$
for $j < i$ and $\alpha \prec_{\varphi_i} \gamma$,
which implies that 
$\alpha \prec_{\Psi} \gamma$.
If $l < i$ holds, we have that 
$\alpha \approx_{\varphi_j} \gamma$ for all $1 \leq j < l$ and  
$\alpha \prec_{\varphi_i} \gamma$. Thus, 
	$\alpha \prec_{\Psi} \gamma$.

\item \emph{Inductive Step:}

\begin{itemize}
\item $\Psi = \Psi_1 \& \Psi_2$. Since
$\alpha \prec_{\Psi} \beta$ then we have 
	$\alpha \prec_{\Psi_1} \beta$ and 
	$\alpha \prec_{\Psi_2} \beta$. Furthermore, 
	$\beta \prec_{\Psi} \gamma$ implies
	$\beta \prec_{\Psi_1} \gamma$ and 
	$\beta \prec_{\Psi_2} \gamma$. From the induction
	hypothesis we have
	$\alpha \prec_{\Psi_1} \gamma$ and 
	$\alpha \prec_{\Psi_2} \gamma$, which leads to
	$\alpha \prec_{\Psi} \gamma$.

\item $\Psi = \Psi_1 \mid \Psi_2$. From 
	$\alpha \prec_{\Psi} \beta$ we have three possible
	cases:
	\begin{enumerate}
	\item $\alpha \prec_{\Psi_1} \beta$ and 
		$\alpha \prec_{\Psi_2} \beta$. The proof for this 
		case is similar to the case

In this case,
		since $\beta \prec_{\Psi} \gamma$ implies
		$\beta \prec_{\Psi_1} \gamma$ and 
		$\beta \prec_{\Psi_2} \gamma$, then we 
		have $\alpha \prec_{\Psi_1} \gamma$ and
		$\alpha \prec_{\Psi_2} \gamma$. This implies that
		$\alpha \prec_{\Psi} \gamma$.

	\item $\alpha \prec_{\Psi_1} \beta$ and 
		$\alpha \approx_{\Psi_2} \beta$. From $\beta \approx_{\Psi}
		\gamma$ we obtain $\beta \approx_{\Psi_1} \gamma$ and
		$\beta \approx_{\Psi_2} \gamma$. Since $\approx_{\Psi_2}$
		is an equivalence relation, we can infer
		$\alpha \approx_{\Psi_2} \gamma$. Furthermore, from the
		induction hypothesis we obtain $\alpha \prec_{\Psi_1} \gamma$.
		These two conclusions lead to $\alpha \prec_{\Psi}
		\gamma$.

	\item $\alpha \prec_{\Psi_2} \beta$ and 
		$\alpha \approx_{\Psi_1} \beta$. This case is 
			symmetrical to the previous one.
	\end{enumerate}

\item $\Psi = ! \Psi_1$. From 
$\alpha \prec_{\Psi} \beta$ we obtain $\beta \prec_{\Psi_1} \alpha$.
Since $\beta \approx_{\Psi} \gamma$ implies 
	$\beta \approx_{\Psi_1} \gamma$, from the induction hypothesis
	we can conclude $\gamma \prec_{\Psi_1} \alpha$ and ultimately
	$\alpha \prec_{\Psi} \gamma$.

\item $\Psi = \Psi_1 \lhd \cdots \lhd \Psi_k$. From the 
definition of $\alpha \prec_{\Psi} \beta$ we have that
there exists $1 \leq i \leq k$ such that
\begin{itemize}
\item $\alpha \approx_{\Psi_j} \beta$ for all $1 \leq j < i$ and
\item $\alpha \prec_{\Psi_i} \beta$
\end{itemize}
Furthermore, from $\beta \approx_{\Psi} \gamma$ we know that
	$\beta \approx_{\Psi_j} \gamma$ for all $1 \leq j \leq k$.
Since all $\approx_{\Psi_j}$ are equivalence relations, then we have
that $\alpha \approx_{\Psi_j} \gamma$ for all $1 \leq j < i$. 
Furthermore, from the induction hypothesis, from 
$\alpha \prec_{\Psi_i} \beta$ and $\beta \approx_{\Psi_i} \gamma$ we
can conclude $\alpha \prec_{\Psi_i} \gamma$. This finally leads to
$\alpha \prec_{\Psi} \gamma$.

\end{itemize}
\end{itemize}
\end{proof}

We now show that 
$\preceq_\Psi$ is a partial order on the set of representatives
of the  equivalence 
classes of $\approx_\Psi$.

\begin{proposition}
\label{p4}
Let $\Psi$ be a general preference. Then 
$\preceq_\Psi$ is a partial order on the set of representatives
of the  equivalence 
classes of $\approx_{\Psi}$.
\end{proposition}
\begin{proof}
We need to show that $\preceq_\Psi$ is reflective, antisymmetric,
and transitive. Reflexivity follows from the fact that 
$\approx_\Phi$ is an equivalence relation and thus
$\alpha \preceq_{\Psi} \alpha$ holds for every $\alpha$.
We prove that $\preceq_{\Psi}$ is antisymmetric
and transitive using structural induction on $\Psi$.

\begin{itemize}

\item \emph{Base:} This corresponds to $\Psi$ being an atomic preference. 
The two properties follow from Proposition \ref{p2}.

\item \emph{Inductive step:} 
Let us consider the possible cases for $\Psi$.
\begin{enumerate}
\item $\Psi = \Psi_1 \& \Psi_2$.

\begin{enumerate}

\item \emph{Anti-symmetry:} consider two representatives
	$\alpha, \beta$ and let us assume that 
	$\alpha \preceq_{\Psi} \beta$ and $\beta \preceq_{\Psi} \alpha$.
	Again, it is enough if we can show that 
	$\alpha \approx_{\Psi} \beta$. 
	$\alpha \preceq_{\Psi} \beta$ implies that 
	$\alpha \preceq_{\Psi_1} \beta$ and 
	$\alpha \preceq_{\Psi_2} \beta$ (Lemma \ref{l2}). 
	$\beta \preceq_{\Psi} \alpha$
	implies that $\beta \preceq_{\Psi_1} \alpha$ and 
	$\beta \preceq_{\Psi_2} \alpha$ (Lemma \ref{l2}). 
	By the induction hypothesis, we have
	that $\alpha \approx_{\Psi_1} \beta$ and 
	$\alpha \approx_{\Psi_2} \beta$ which imply that 
	$\alpha \approx_{\Psi} \beta$.

\item \emph{Transitivity:} consider three representatives 
	$\alpha_1$, $\alpha_2$, and $\alpha_3$ with 
	$\alpha_1 \preceq_{\Psi} \alpha_2$ and 
	$\alpha_2 \preceq_{\Psi} \alpha_3$.
	The first relationship implies that 
	$\alpha_1 \preceq_{\Psi_1} \alpha_2$ and 
	$\alpha_1 \preceq_{\Psi_2} \alpha_2$. The second 
	relationship implies that 
	$\alpha_2 \preceq_{\Psi_1} \alpha_3$, and
	$\alpha_2 \preceq_{\Psi_2} \alpha_3$.
	From the induction hypothesis we have 
	$\alpha_1 \preceq_{\Psi_1} \alpha_3$ and
	$\alpha_1 \preceq_{\Psi_2} \alpha_3$. Thus, 
	$\alpha_1 \preceq_{\Psi} \alpha_3$.
\end{enumerate}

\item $\Psi = \Psi_1 \mid \Psi_2$. Similar arguments 
	to the above case (with the help of Lemma \ref{l2})
	allows us to conclude that anti-symmetry and 
	transitivity also holds for this case.

\item $\Psi = \:! \Psi_1$. 
	
\begin{enumerate}
\item \emph{Anti-symmetry:} consider two representatives
	$\alpha, \beta$ and let us assume that 
	$\alpha \preceq_{\Psi} \beta$ and $\beta \preceq_{\Psi} \alpha$.
	Lemma \ref{l2} imply that 
	$\beta \preceq_{\Psi_1} \alpha$ and 
	$\alpha \preceq_{\Psi_1} \beta$. 
	By the induction hypothesis, we have
	that $\alpha \approx_{\Psi_1} \beta$. This allows us 
	to conclude that $\alpha \approx_{\Psi} \beta$.

\item \emph{Transitivity:} consider three representatives 
	$\alpha_1$, $\alpha_2$, and $\alpha_3$ with 
	$\alpha_1 \preceq_{\Psi} \alpha_2$ and 
	$\alpha_2 \preceq_{\Psi} \alpha_3$.
	Again, Lemma \ref{l2} implies that 
	$\alpha_2 \preceq_{\Psi_1} \alpha_1$ and 
	$\alpha_3 \preceq_{\Psi_1} \alpha_2$. 
	The induction hypothesis implies that
	$\alpha_3 \preceq_{\Psi_1} \alpha_1$, and hence, 
	$\alpha_1 \preceq_{\Psi} \alpha_3$.
\end{enumerate}

\item $\Psi= \Psi_1 \lhd \cdots \lhd \Psi_k$.

\begin{enumerate}
\item \emph{Anti-symmetry:} consider two representatives
	$\alpha, \beta$ with 
	$\alpha \preceq_{\Psi} \beta$ and $\beta \preceq_{\Psi} \alpha$.
	Assume that $\alpha \prec_{\Psi} \beta$. 
	This means that there exists $1 \le i \le k$ such that 
	$\alpha \approx_{\Psi_j} \beta$ for all $1 \le j < i$ and 
	$\alpha \prec_{\Psi_i} \beta$. This will imply that 
	$\beta \preceq_{\Psi} \alpha$ cannot hold, i.e., we have 
	a contradiction. This means that 
	$\alpha \approx_{\Psi} \beta$. 

\item \emph{Transitivity:} consider three representatives 
	$\alpha_1$, $\alpha_2$, and $\alpha_3$ with 
	$\alpha_1 \preceq_{\Psi} \alpha_2$ and 
	$\alpha_2 \preceq_{\Psi} \alpha_3$. We have 
	four sub-cases:
\begin{enumerate}
\item $\alpha_1 \approx_{\Psi} \alpha_2$ and $\alpha_2 \approx_{\Psi} \alpha_3$.
In this case, we have that 
$\alpha_1 \approx_{\Psi} \alpha_3$ because 
$\approx_\Psi$ is an equivalence relation.

\item $\alpha_1 \prec_{\Psi} \alpha_2$ and $\alpha_2 \approx_{\Psi} \alpha_3$.
Lemma \ref{sideq} implies that 
$\alpha_1 \prec_{\Psi} \alpha_3$.

\item $\alpha_1 \approx_{\Psi} \alpha_2$ and $\alpha_2 \prec_{\Psi} \alpha_3$.
Lemma \ref{sideq} implies that 
$\alpha_1 \prec_{\Psi} \alpha_3$.

\item $\alpha_1 \prec_{\Psi} \alpha_2$ and $\alpha_2 \prec_{\Psi} \alpha_3$.
	This implies that 
	there exists $1 \leq i_1, i_2 \leq k$ such that
 	$\alpha_1 \approx_{\Psi_j} \alpha_2$ for all $1 \le j < i_1$ and 
	$\alpha_1 \prec_{\Psi_{i_1}} \alpha_2$; 
	and 
	$\alpha_2 \approx_{\Psi_j} \alpha_3$ for all $1 \le j < i_2$ and 
	$\alpha_2 \prec_{\Psi_{i_2}} \alpha_3$. 	

	If $i_1 < i_2$ then from the fact that $\approx_{\Psi_j}$ are
	equivalence relations, we can conclude that 
	$\alpha_1 \approx_{\Psi_j} \alpha_3$ for all $1 \leq j < i_1$. 
	Furthermore, by Lemma \ref{sideq}, from 
	$\alpha_1 \prec_{\Psi_{i_1}} \alpha_2$ and 
	$\alpha_2 \approx_{\Psi_{i_1}} \alpha_3$, we can conclude that
	$\alpha_1 \prec_{\Psi_{i_1}} \alpha_3$. This leads to 
	$\alpha_1 \prec_{\Psi} \alpha_3$.

	Similarly, if $i_1 > i_2$, we have that 
	$\alpha_1 \approx_{\Psi_j} \alpha_3$ for all $1 \leq j < i_2$
	and $\alpha_1 \prec_{\Psi_{i_2}} \alpha_3$.	
	This leads to $\alpha_1 \prec_{\Psi} \alpha_3$.

	Finally, if $i_1=i_2$, then we have that  
	$\alpha_1 \approx_{\Psi_j} \alpha_3$ 
	for all $1 \leq j < i_1$. Furthermore, 
	 since $\alpha_1 \prec_{\Psi_i} \alpha_2$ and 
			$\alpha_2 \prec_{\Psi_i} \alpha_3$, from the 
		induction hypothesis we obtain
		$\alpha_1 \prec_{\Psi_i} \alpha_3$.
	This also leads to $\alpha_1 \prec_{\Psi} \alpha_3$.
\end{enumerate}
\end{enumerate}
\end{enumerate}
\end{itemize}

\end{proof}

\begin{definition}
Given a general preference $\Psi$, we say that 
a trajectory $\alpha$ is {\em most preferred} if there is no 
 trajectory that is preferred to $\alpha$.
\end{definition}
The next example highlights some differences 
and similarities between basic desires and general preferences.
\begin{example}
\label{ex5}
Let us consider the original action theory presented in Section \ref{sec2}
with the action {\em buy\_coffee} and a user having the goal of being at the
school and having coffee. Intuitively, every trajectory achieving the goal 
of the user would require the action of going to the coffee shop, buying 
the coffee, and going to the school thereafter.

Let us consider the following two preferences
(similar to those discussed in Example \ref{ex4}):
\[
time = \always(occ(buy\_coffee) \vee (take\_taxi <^e (drive \vee bus) <^e walk)) 
\]
and  
\[
cost = \always(occ(buy\_coffee) \vee (walk <^e (drive \vee bus) <^e take\_taxi)).  
\]
It is easy to see that most preferred trajectories with respect to 
{\em time} should contain only the actions {\em buy\_coffee} 
and {\em take\_taxi} while most preferred trajectories with respect to 
{\em cost} should contain only the actions {\em buy\_coffee} 
and {\em walk}. 

Consider the two preferences:
\[
\Psi_1 = time \wedge cost
\]
and 
\[
\Psi_2 = time \; \& \; cost.
\]
Observe that $\Psi_1$ is a basic desire while $\Psi_2$ is a general 
preference. 
It is easy to see that there are no trajectories satisfying 
the preference $\Psi_1$. Thus, every trajectory achieving the goal 
is a most preferred trajectory w.r.t. $\Psi_1$. At the same time, we can show 
that for every pair of trajectories  $\alpha$ and $\beta$,
neither $\alpha \prec_{\Psi_2} \beta$ nor 
$\beta \prec_{\Psi_2} \alpha$ holds.

Let us now consider the two preferences:
\[
\Psi_3 = time \vee cost
\]
and 
\[
\Psi_4 = time \mid cost
\]
with respect to the same set of trajectories.
Here, $\Psi_3$ is a basic desire while $\Psi_4$ is a general preference. 
We can see that any trajectory containing the actions 
$taxi$ and $walk$ would be most preferred with respect to 
$\Psi_3$. All of these trajectories are indistinguishable.
For example, the trajectory 
\[
\alpha = s_0\; walk(home,coffee\_shop) \; s_1 \; buy\_coffee \; s_2 \; walk(coffee\_shop, school)\; s_3
\]
and the trajectory 
\[
\beta = s_0 \; walk(home,coffee\_shop)\; s_1'\; buy\_coffee \; s'_2 \; 
take\_taxi(coffee\_shop, school)\; s_3'
\] 
are indistinguishable with respect to $\Psi_3$. 
On the other hand, we have that 
$\alpha \prec_{cost} \beta$ (the minimal cost action is always
used) and 
$\alpha \approx_{time} \beta$ 
(the fastest action is not used every time). This implies
that $\alpha \prec_{\Psi_4} \beta$.

Let us consider now the preference $$\Psi_5 = !\: time.$$ It is easy to see
that $\Psi_5$ is equivalent to $cost$ in the sense that every most 
preferred trajectory w.r.t. $\Psi_5$ is a most 
preferred trajectory w.r.t {\em cost} and vice versa. 
\hfill$\Box$
\end{example}
The next proposition is of interest for the computation of 
preferred trajectories\footnote{
   We would like to thank the anonymous reviewer who pointed out 
   that this proposition is necessary for the computation presented 
   in the next section.
}.

\begin{proposition} \label{associative}
Let $\Psi_1, \; \Psi_2$ and $\Psi_3$ be three 
general preferences, 
$\Psi = \Psi_1 \lhd \Psi_2 \lhd \Psi_3$, 
and $\Gamma = \Psi_1 \lhd (\Psi_2 \lhd \Psi_3)$.
For arbitrary trajectories $\alpha$ and $\beta$, the following holds:
\begin{itemize}
\item $\alpha \approx_\Psi \beta$ if and only if 
	$\alpha \approx_\Gamma \beta$; and 

\item $\alpha \prec_\Psi \beta$ if and only if 
	$\alpha \prec_\Gamma \beta$.
\end{itemize}
\end{proposition}
\begin{proof}
\begin{itemize}
\item
We have that 
$\alpha \approx_\Psi \beta$ iff 
$\alpha \approx_{\Psi_i} \beta$ for $i \in \{1,2,3\}$ iff
$\alpha \approx_{\Psi_1} \beta$ 
and 
$\alpha \approx_{\Psi_2 \lhd \Psi_3} \beta$ iff
$\alpha \approx_\Gamma \beta$.

\item Since 
$\alpha \prec_\Psi \beta$ iff 
there exists $i \in \{1,2,3\}$ such that 
$\alpha \approx_{\Psi_j} \beta$ for $1 \le j < i$ 
and 
$\alpha \prec_{\Psi_i} \beta$, we have three cases: $i=1$, 2, or 3.
We consider these three cases:

\begin{itemize}
\item $i = 1$. This implies immediately that 
$\alpha \prec_\Gamma \beta$.

\item $i = 2$. This means that 
$\alpha \approx_{\Psi_1} \beta$ and 
$\alpha \prec_{\Psi_2} \beta$. This in turn 
implies that 
$\alpha \approx_{\Psi_1} \beta$ and 
$\alpha \prec_{\Psi_2 \lhd \Psi_3} \beta$, i.e.,
$\alpha \prec_\Gamma \beta$.

\item $i=3$. This is similar to the case $i=2$.
\end{itemize}
Thus, we have that if $\alpha \prec_\Psi \beta$ 
then 
$\alpha \prec_\Gamma \beta$. The proof of the reverse 
is similar.
\end{itemize}
\end{proof}

\section{Computing Preferred Trajectories}
\label{sec4}\label{implementation}

In this section, we address the problem of computing preferred 
trajectories. The ability to use the operators
$\wedge, \neg, \vee$ as 
well as $\&, |, !$ 
in the construction of preference formulae allows us
to combine several preferences into a preference formula.
For example, if a user has two atomic preferences $\Psi$ and $\Phi$,
but does not prefer $\Psi$ over $\Phi$ or vice versa,
he can combine them in to a single preference
$\Psi \wedge \Phi  \lhd  \Psi \vee \Phi  \lhd \neg \Psi \wedge \neg \Phi$.
The same can be done if $\Psi$ or $\Phi$ are general preferences. 
Thus, without loss of generality, we can assume that we only have 
one preference formula to deal with.
We would like to note that this way of combination of 
   preferences creates a preference whose size could be exponential 
   in the number of preferences. We believe, however, 
   that it is more likely that a user---when presented with 
   a set of preferences---will have a preferred order 
   on some of these preferences. This information can be used to 
   create a single preference with a reasonable and manageable size. 

Given a planning problem 
$\langle D,I,G \rangle$ and a preference formula $\varphi$, 
we are interested in finding 
a most preferred trajectory $\alpha$ achieving $G$ 
w.r.t. the preference $\varphi$. 
We will show how this can be done in answer set programming.

A naive encoding could be realized by modeling 
$\langle D,I,G \rangle$ in logic programming (following the
scheme described in \cite{sbm02a}), using an answer set 
engine to determine its answer sets---and, thus, the
trajectories---and then filtering them according to $\varphi$.

In the rest of this section, instead, we present a more sophisticated
approach which allows us to determine \emph{a} most preferred
trajectory.
We achieve this by encoding each basic desire $\varphi$
as a set of rules $\Pi_\varphi$ and by developing two sets of 
rules $\Pi_{sat}$ and $\Pi_{pref}$. The program 
$\Pi_{sat}$ checks whether 
a basic desire is satisfied by a trajectory. On the other hand,
$\Pi_{pref}$ consists of rules that, when used together with the {\bf maximize} 
construct of {\bf smodels}, allow us to find a most 
preferred trajectory with respect to a preference formula.
Since $\Pi(D,I,G)$ has already been discussed in Section \ref{sec2}, 
we will start by
defining $\Pi_\varphi$. 

\subsection{$\Pi_\varphi$: Encoding of Basic Desire Formulae}
\label{enconde-desire}

The encoding of a basic desire formula is similar to the encoding 
of a fluent formula proposed in \cite{sbm02a}. In 
our encoding, we will use the predicate $desire$ as a domain 
predicate. The elements of the set 
$\{desire(l) \mid  l$ is a fluent literal$\}$
belong to $\Pi_\varphi$. Each atom in this set 
declares the fact that each literal is also a desire. Next,
we associate to each basic desire 
formula $\varphi$ a unique name 
$n_\varphi$. If $\varphi$ is a basic desire formula then 
it will be encoded as a set of facts, denoted by $\Pi_\varphi$. 
This set is defined inductively over the structure 
of $\varphi$\footnote{To simplify this encoding, we have developed a
Prolog program that translates $\varphi$ into $\Pi_\varphi$.
This program can be downloaded from 
\url{http://www.cs.nmsu.edu/~tson/ASPlan/Preferences/conv_form.pl}.
}.
The encoding is performed as follows.
\begin{list}{$-$}{\topsep=2pt \itemsep=1pt \parsep=0pt}
\item If $\varphi$ is a fluent literal $l$ then $\Pi_\varphi = \{desire(l)\}$;
\item If $\varphi = goal(\phi)$ then $\Pi_\varphi = \Pi_\phi \cup 
	\{desire(n_\varphi), goal(n_\phi)\}$;
\item If $\varphi = occ(a)$ then $\Pi_\varphi = \{desire(n_\varphi), 
	happen(n_\varphi, a)\}$;
\item If $\varphi = \varphi_1 \wedge \varphi_2$ 
      then $\Pi_\varphi = \Pi_{\varphi_1} \cup \Pi_{\varphi_2} \cup 
	\{desire(n_\varphi), and(n_\varphi, n_{\varphi_1},n_{\varphi_2})\}$;
\item If $\varphi = \varphi_1 \vee \varphi_2$ 
      then $\Pi_\varphi = \Pi_{\varphi_1} \cup \Pi_{\varphi_2} \cup 
	\{desire(n_\varphi), or(n_\varphi, n_{\varphi_1},n_{\varphi_2})\}$;
\item If $\varphi = \neg \phi$ 
      then $\Pi_\varphi = \Pi_\phi \cup 	
	\{desire(n_\varphi), negation(n_\varphi, n_\phi)\}$;
\item If $\varphi = \next(\phi)$ 
      then $\Pi_\varphi = \Pi_\phi \cup 
     \{desire(n_\varphi), next(n_\varphi, n_\phi)\}$;
\item If $\varphi = \until(\varphi_1,\varphi_2)$ 
      then $\Pi_\varphi = \Pi_{\varphi_1} \cup \Pi_{\varphi_2} \cup 
     \{desire(n_\varphi), until(n_\varphi, n_{\varphi_1},n_{\varphi_2})\}$;
\item If $\varphi = \always(\phi)$ 
      then $\Pi_\varphi = \Pi_\phi \cup 
     \{desire(n_\varphi), always(n_\varphi, n_\phi)\}$;
\item If $\varphi = \eventually(\phi)$ 
      then $\Pi_\varphi = \Pi_\phi \cup 
     \{desire(n_\varphi), eventually(n_\varphi, n_\phi)\}$.
\end{list}
It is worth noting that, due to the uniqueness of names for basic desires, 
$n_\varphi$ will not occur in $\Pi_{\Phi} \setminus \{desire(n_\varphi)\}$.

\subsection{$\Pi_{sat}$: Rules for Checking of Basic Desire Formula Satisfaction}

We now present the set of rules that checks whether a trajectory
satisfies a basic desire formula. Recall that an answer set
of the program $\Pi(D,I,G)$ will contain a trajectory 
where action occurrences are recorded by atoms of the form 
$occ(a,t)$ and the truth value of fluent literals 
is represented by atoms of the form $holds(f,t)$, 
where $a \in \mathbf{A}$, $f$ is a fluent literal,
and $t$ is a time moment between $0$ and $length$. The program
$\Pi_{sat}$ defines the predicate 
$satisfy(F,T)$, where $F$ and $T$ are variables representing 
a basic desire and a time moment, respectively.
The satisfiability of a fluent formula at a time moment 
will be defined by the predicate $h(F,T)$---which 
builds on the previous mentioned predicate $holds$ and the usual 
logical operators, such as $\wedge, \vee, \neg$. 
Intuitively, $satisfy(F,T)$ says that the basic desire 
$F$ is satisfied by the trajectory starting from the 
time moment $T$.  The rules of $\Pi_{sat}$ are defined inductively
on the structure of $F$ and are given next.
\begin{eqnarray}
satisfy(F, T) & \leftarrow & desire(F),
			     goal(F),
			     satisfy(F, length). \label{rtk0} \\
satisfy(F, T) & \leftarrow & desire(F),
			     happen(F, A),
			     occ(A, T). \label{rtk1} \\
satisfy(F, T) & \leftarrow & desire(F),
			     literal(F),
			     holds(F, T). \label{rtk2} \\
satisfy(F, T) & \leftarrow & desire(F),
			     and(F, F_1, F_2),   \label{rtk3} \\
              &   & satisfy(F_1,T),  satisfy(F_2,T). \nonumber \\
satisfy(F, T) & \leftarrow & desire(F),
			     or(F, F_1, F_2),  satisfy(F_1,T).\label{rtk3or1} 
\\
satisfy(F, T) & \leftarrow & desire(F),
			     or(F, F_1, F_2),  satisfy(F_2,T).\label{rtk3or2} 
\\
satisfy(F, T) & \la & desire(F), 
                      negation(F, F_1),  \naf satisfy(F_1, T).\;\;\;\;\;\;\;\;\ \label{rtk12} \\
satisfy(F, T) & \la & desire(F), 
		      until(F, F_1, F_2), T < T_1,   \label{rtk6case1} \\
	      &	    &	during(F_1, T, T_1-1), satisfy(F_2, T_1). \nonumber \\
satisfy(F, T) & \la & desire(F), 
		      until(F, F_1, F_2), satisfy(F_2, T).   \label{rtk6case2} \\
satisfy(F, T) & \la & desire(F), 
		      always(F, F_1),  during(F_1, T, length).  \label{rtk7} \\
satisfy(F, T) & \la & desire(F), 
                      next(F, F_1), satisfy(F_1, T+1).  \label{rtk9} \\
satisfy(F, T) & \la & desire(F), 
		      eventually(F, F_1), T \le T1, \label{rtk13} \\ 
		& & 	satisfy(F_1, T1). \nonumber \\ 
during(F, T, T_1) & \la & T < T_1,
                      desire(F),
                      satisfy(F, T),\label{rtk10} \\
                    &     & during(F, T+1, T_1). \nonumber \\
during(F, T, T) & \la & desire(F),
                      satisfy(F, T). \label{rtk11} 
\end{eqnarray}
In the next theorem, we prove the correctness of $\Pi_{sat}$. 
We need some additional notation. 
For a trajectory $\alpha=s_0a_1\ldots a_{n}s_n$,
let 
\[\alpha^{-1} = \{occ(a_i,i-1) \mid i \in \{1,\ldots,n\}\} \cup
\{holds(f,i) \mid f \in s_i, i \in \{0,\ldots,n\}\}.\]
 We have:
\begin{theorem} \label{th-basic-desire}
Let $\langle D,I,G \rangle$ be a planning problem,
$\alpha=s_0a_1\ldots a_{n}s_n$ be a trajectory, and $\varphi$ 
be a basic desire formula. 
Then, for every $t$, $0 \le t \le length$ and
every basic desire formula $\eta$ with $desire(n_\eta) \in \Pi_\varphi$, 
\[
\alpha[t] \models \eta
\;\;\;\;\; 
\textnormal{ iff } 
\;\;\;\;\; 
\Pi_\varphi \cup \Pi_{sat} \cup (\alpha)^{-1} \models 
 satisfy(n_\eta,t).
\]
(Recall that $\alpha[t]$ is the trajectory $s_ta_{t+1}\ldots a_{n}s_n$.)
\end{theorem}
\begin{proof}
First, we prove that the program 
$\Pi = \Pi_\varphi \cup \Pi_{sat} \cup (\alpha)^{-1}$ 
has only one answer set. It is well-known that if a program is locally
stratified then it has a unique answer set \cite{apt88,prz88b}. We
will show that $\Pi$ (more precisely, the set of ground instances of rules in
$\Pi$) is indeed locally stratified. To accomplish this we need to
find a mapping $\lambda$ from literals of the grounding of 
$\Pi$ to $\mathbf{N}$
that has the property: if 
$$A_0 \leftarrow A_1, A_2, \ldots A_n,
\naf B_1, \naf B_2, \ldots \naf B_m$$
is a rule in $\Pi$, then
$\lambda(A_0) \geq \lambda(A_i)$ for all $1\leq i\leq n$ and
$\lambda(A_0) > \lambda(B_j)$ for all $1\leq j\leq m$.

To define $\lambda$, we first associate a non-negative number
$\sigma(\phi)$ to each constant $n_\eta$ as follows. Intuitively, 
$\sigma(\phi)$ represents the complexity of $\phi$.
\begin{itemize}
\item $\sigma(l) = 0$ if $l$ is a literal.

\item $\sigma(n_\eta) = 0$ if $\eta$ has the form $occ(a)$.

\item $\sigma(n_\eta) = \sigma(n_{\eta_1}) + 1$
if $\eta$ has the form $\neg \eta_1$,
$\next(\eta_1)$,  $\eventually(\eta_1)$, or $\always(\eta_1)$.

\item $\sigma(n_\eta) = \max \{\sigma(n_{\eta_1}), \sigma(n_{\eta_2})\} + 1$
if $\eta$ has the form $\eta_1 \wedge \eta_2$,
$\eta_1 \vee \eta_2$, or $\until(\eta_1,\eta_2)$.

\item $\sigma(n_\eta) = \sigma(n_{\eta_1})$ if $\eta = \goal(\eta_1)$. 
\end{itemize}

\ni
We define $\lambda$ as follows. 
\begin{itemize}
\item $\lambda(satisfy(n_\eta,t)) = 5 * \sigma(\eta) + 2$,
\item $\lambda(during(n_\eta,t,t')) = 5 * \sigma(\eta) + 4$, and
\item $\lambda(l) = 0$ for every other literal of $\Pi$.
\end{itemize}

\ni
We need to check that $\lambda$ satisfies the necessary requirements. 
For example, for the rule (\ref{rtk0}), we have that 
$\lambda(satisfy(n_{\eta},t)) = \lambda(satisfy(n_\eta, length)) = 
5*\sigma(n_\eta) + 2$
and $\lambda(satisfy(n_\eta,t)) \ge 2 > 0 = \lambda(goal(n_\eta)) = 
	\lambda(desire(n_\eta))$.
For the rule (\ref{rtk12}), we have that 
$\lambda(satisfy(n_\eta,t)) = 5*\sigma(n_\eta) + 2 = 
5*(\sigma(n_{\eta_1})+1)+2 > 
5*(\sigma(n_{\eta_1}))+2 = \lambda(satisfy(n_{\eta_1}, length))$. 
The verification of this property for other rules is similar. 
Thus, we can conclude that 
$\Pi$ has only one answer set. 
Let $X$ be the  answer set of $\Pi$.
We prove the conclusion of the theorem by induction over $\sigma(n_\eta)$.

\st {\bf Base:} Let $\eta$ be a formula with $\sigma(n_\eta) = 0$.
By the definition of $\sigma$, we know that $\eta$ is a fluent literal 
or of the form $occ(a)$. If $\eta$ is a literal,
then $\eta$ is true in $s_t$ iff $\eta$ is in $s_t$, that is, iff
$holds(\eta,t)$ belongs to $X$, which, because of
rule (\ref{rtk2}), proves the base case. If $\eta = occ(a)$
then we know that $happen(n_\eta,a) \in \Pi_\varphi$.
Thus, $satisfy(n_\eta, t) \in X$ iff $occ(a, t) \in X$ (because 
of the rule (\ref{rtk1})) iff $\alpha[t] \models \eta$.

\st {\bf Step:} Assume that for all $0\leq j\leq k$ and formula
$\eta$ such that $\sigma(n_\eta) = j$, $\alpha[t] \models \eta$
iff $satisfy(n_\eta,t)$ is in $X$.

\st
Let $\eta$ be such a formula that $\sigma(n_\eta) = k + 1$.
Because of the definition of $\sigma$, 
we have the following cases:
\begin{itemize}
\item {\bf Case 1:} $\eta = \neg \eta_1$.
We have that $\sigma(n_{\eta_1}) = \sigma(n_\eta) - 1 = k$.
Since $negation(n_\eta, n_{\eta_1}) \in X$,
$satisfy(n_\eta,t) \in X$ iff the body of
rule (\ref{rtk12}) is satisfied by $X$ iff
$satisfy(n_{\eta_1}, t) \notin X$ iff
$\alpha[t] \not\models \eta_1$ (by the induction hypothesis) iff
$\alpha[t] \models \eta$.
\item {\bf Case 2:} $\eta = \eta_1 \wedge \eta_2$.
Similar to the first case, it follows from the rule (\ref{rtk3})
and the facts $desire(n_\eta)$ and $and(n_\eta, n_{\eta_1}, n_{\eta_2})$ 
in $\Pi_\varphi$ that
$satisfy(n_\eta,t) \in X$ iff the body of
rule (\ref{rtk3}) is satisfied by $X$ iff
$satisfy(n_{\eta_1}, t) \in X$ and $satisfy(n_{\eta_2}, t) \in X$
iff $\alpha[t] \models \eta_1$ and $\alpha[t] \models \eta_2$ 
(induction hypothesis)
iff $\alpha[t] \models \eta$.
\item {\bf Case 3:} $\eta = \eta_1 \vee \eta_2$.
The proof is similar to the above cases, relying on the
two rules (\ref{rtk3or1}), (\ref{rtk3or2}), and
the facts that $desire(n_\eta) \in \Pi_\varphi$ 
and $or(n_\eta, n_{\eta_1}, n_{\eta_2}) \in \Pi_\varphi$.

\item {\bf Case 4:} $\eta = \until(\eta_1,\eta_2)$.
We have that $\sigma(n_{\eta_1})\leq k$ and $\sigma(n_{\eta_2})\leq k$.
Assume that $\alpha[t] \models \eta$. By Definition \ref{def4},
there exists $t \leq t_2 \leq n$ such that $\alpha[t_2] \models
\eta_2$ and for all $t \leq t_1 < t_2$, $\alpha[t_1] \models \eta_1$.
By the induction hypothesis, $satisfy(n_{\eta_2},t_2) \in X$ and 
$satisfy(n_{\eta_1},t_1) \in X$ 
for $t \leq t_1 < t_2$. It follows
that $during(n_{\eta_1},t,t_2-1) \in X$.
Because of rule (\ref{rtk6case1})-(\ref{rtk6case2}), we have
$satisfy(n_\eta,t) \in X$. 

On the other hand, if $satisfy(n_\eta,t) \in X$,
because the only rules supporting $satisfy(n_\eta,t)$ are 
(\ref{rtk6case1})-(\ref{rtk6case2}), 
there exists $t_2$, $t \leq t_2 \leq length$, and
$during(n_{\eta_1},t,t_2-1) \in X$, and $satisfy(n_{\eta_2},t_2)$. It
follows from $during(n_{\eta_1},t,t_2-1) \in X$ that
$satisfy(n_{\eta_1},t_1) \in X$ for all $t \leq t_1 < t_2$. By
the induction hypothesis, we have $\alpha[t_1] \models \eta_1$ for all $t
\leq t_1 < t_2$ and $\alpha[t_2] \models \eta_2$. Thus $\alpha[t] \models
\until(\eta_1,\eta_2)$, i.e., $\alpha[t] \models \eta$.
\item {\bf Case 5:} $\eta = \next(\eta_1)$. Note that 
$\sigma(n_{\eta_1})\leq k$.
Rule (\ref{rtk9}) is the only rule supporting $satisfy(n_\eta,t)$
where $\eta = \next(\eta_1)$. So
$satisfy(n_\eta,t) \in X$ iff  $satisfy(n_{\eta_1},t+1) \in X$
iff $\alpha[t+1] \models \eta_1$
iff $\alpha[t] \models \next(\eta_1)$.
\item {\bf Case 6:} $\eta = \always(\eta_1)$.
We note that $\sigma(n_{\eta_1}) \leq k$.
Observe that $satisfy(n_\eta,t)$ is supported only by rule (\ref{rtk7}).
So $satisfy(n_\eta,t) \in X$ iff $during(n_{\eta_1},t,n) \in X$.
The latter happens iff $satisfy(n_{\eta_1},t_1) \in X$ for all $t \leq t_1 \leq n$,
that is, iff $\alpha[t_1] \models \eta_1$ for all $t \leq t_1 \leq n$ which is
equivalent to $\alpha[t] \models \always(\eta_1)$, i.e., iff 
$\alpha[t] \models \eta$.
\item {\bf Case 7:} $\eta = \eventually(\eta_1)$.
We know that $satisfy(n_\eta,t)\in X$ is supported only by
rule (\ref{rtk13}). So $satisfy(n_\eta,t) \in X$ iff there exists
$t \leq t_1 \leq n$ such that $satisfy(n_{\eta_1},t_1) \in X$.
Because $\sigma(n_{\eta_1}) \leq k$, by induction, $satisfy(n_\eta,t) \in X$
iff there exists $t \leq t_1 \leq n$ such that
$\alpha[t_1] \models \eta_1$, that is, 
iff $\alpha[t] \models \eventually(\eta_1)$,
i.e., iff $\alpha[t] \models \eta$.

\item {\bf Case 8:} $\eta = \goal(\eta_1)$. 
Since $\eta_1$ does not contain the $\goal$ operator, 
it follows from the above cases that 
$satisfy(n_{\eta_1}, n) \in X$ iff 
$\alpha[n] \models \eta_1$. From the rule (\ref{rtk0}),
we can conclude that 
$satisfy(n_{\eta}, t) \in X$ iff 
$satisfy(n_{\eta_1}, n) \in X$ iff 
$\alpha[t] \models \eta$. 
\end{itemize}
\ni
The above cases prove the inductive step and, hence, the theorem.
\end{proof}
The next theorem follows from Theorems (\ref{th1})
and (\ref{th-basic-desire}).

\begin{theorem} \label{th-basic-desire-compute}
Let $\langle D,I,G \rangle$ be a planning problem and $\varphi$ 
be a basic desire formula. 
For every answer set $M$ of the program 
$\Pi(D,I,G,\varphi) = \Pi(D,I,G) \cup \Pi_\varphi \cup \Pi_{sat}$, 
\[
\alpha_M \models \varphi
\;\;\;\;\; 
\textnormal{ iff } 
\;\;\;\;\; 
 satisfy(n_\varphi,0) \in M.
\]
where $\alpha_M$ denotes the trajectory achieving $G$ 
represented by $M$.
\end{theorem}
\begin{proof}
Let $S$ be the set of literals of the program $\Pi(D,I,G)$.
It is easy to see that for every rule $r$ in $\Pi(D,I,G,\varphi)$ if 
the head of $r$ belongs to $S$ then every literal occurring in the body 
of $r$ also belongs to $S$. Thus, 
$S$ is a splitting set of $\Pi(D,I,G,\varphi)$. Using 
the Splitting Theorem \cite{lif94a}, we can show that 
$M$ is an answer set of 
$\Pi(D,I,G,\varphi)$ iff $M = X \cup Y$, where $X$ is an
answer set of $\Pi(D,I,G)$ and $Y$ is an answer set 
of the program $\Pi_\varphi \cup \Pi_{sat} \cup (\alpha_M)^{-1}$. 
It follows from Theorem \ref{th-basic-desire} that 
$\alpha_M[0] \models \varphi$ iff 
$satisfy(n_\varphi, 0) \in Y$ iff 
$satisfy(n_\varphi, 0) \in M$. 
\end{proof}
The above theorem allows us to compute a most preferred trajectory 
using the {\bf smodels} system. 
Let $\Pi(D,I,G,\varphi)$ be the program consisting of the $\Pi(D,I,G) \cup  
\Pi_\varphi \cup \Pi_{sat}$ and the rule
\begin{equation} \label{rule11}
\mathbf{maximize}\{satisfy(n_\varphi,0) = 1, \naf satisfy(n_\varphi,0) = 0\}.
\end{equation}
Note that rule (\ref{rule11}) represents the fact that
the answer sets in which 
$satisfy(n_\varphi,0)$ holds are most preferred.
{\bf smodels} will first try to compute answer 
sets of $\Pi$ in which $satisfy(n_\varphi,0)$ holds; only if no answer 
sets with this property exist, other answer sets will be 
considered.\footnote{The current implementation 
of {\bf smodels} has some restrictions 
on using the {\bf maximize} construct; our {\bf jsmodels} 
system can now deal with this construct properly.} 
\begin{theorem}
Let $\langle D,I,G \rangle$ be a planning problem and $\varphi$ 
be a basic desire formula. For 
every answer set $M$ of $\Pi(D,I,G,\varphi)$, 
if $satisfy(n_\varphi,0) \in M$ then 
$\alpha_M$ is a most preferred trajectory w.r.t. $\varphi$.
\end{theorem}
\begin{proof}
The result  follows directly from Theorem \ref{th-basic-desire-compute}:
$satisfy(n_\varphi,0) \in M$ implies that 
$\alpha_M \models \varphi$, hence
$\alpha_M$ is a most preferred trajectory w.r.t. $\varphi$.
\end{proof}
The above theorem gives us a way to compute a most preferred 
trajectory with respect to a basic desire. We will now generalize 
this approach to deal with atomic and general preferences. 
The intuition is to associate to the different components
of the preference formula a \emph{weight}; these weights
are then used to obtain a weight for each 
trajectory, based on
what components of the preference formula are satisfied
by the trajectory. The {\bf maximize} construct in {\bf smodels}
can then be used to compute answer sets with maximal weight,
thus computing most preferred trajectories. The
weight functions are defined as follows. 
\begin{definition}
Given a general preference $\Psi$, {\em a weight function $w_\Psi$ 
w.r.t. $\Psi$} (or weight function, for short, when it is 
clear from the context what is $\Psi$) 
assigns to each trajectory $\alpha$ a non-negative 
number $w_\Psi(\alpha)$.
\end{definition}
Since our goal is to use weight functions in generating 
most preferred trajectories using the {\bf maximize} construct 
in {\bf smodels}, we would like to find weight functions
which assign the maximal weight to most preferred trajectories.
In what follows, we discuss a class of weight functions that 
satisfy this property. 
\begin{definition}
Let $\Psi$ be a general preference. A weight function  
$w_\Psi$ is called {\em admissible} if 
it satisfies the following properties:
\begin{itemize}
  \item[]{\bf (i)} if $\alpha \prec_\Psi \beta$ then 
	$w_\Psi(\alpha) > w_\Psi(\beta)$; and
  \item[] {\bf (ii)} if $\alpha \approx_\Psi \beta$ then 
	$w_\Psi(\alpha) = w_\Psi(\beta)$.
\end{itemize}
\end{definition}
It is easy to see that if $w_\Psi$ 
is admissible then the following theorem will hold.
\begin{proposition}
Let $\Psi$ be a general preference formula and 
$w_{\Psi}(\alpha)$ be an admissible weight function. 
If $\alpha$ is 
a trajectory such that $w_{\Psi}(\alpha)$ is maximal, then 
$\alpha$ is a most preferred trajectory w.r.t. $\Psi$.
\end{proposition}
\begin{proof}
Since $w_{\Psi}(\alpha)$ is maximal and $w_\Psi$ is admissible,
we conclude that there exists no trajectory $\beta$ such that 
$\beta \prec_\Psi \alpha$. Thus, $\alpha$ is a most preferred
trajectory w.r.t. $\Psi$. 
\end{proof}
The above proposition implies that we can compute a most preferred 
trajectory using {\bf smodels} if we can implement an admissible 
weight function. This is the topic of the next section. 
We would like to emphasize 
that the above result states \emph{soundness} of this method, but
\emph{not completeness}. This means that the computation scheme proposed in the
next section will return \emph{a} most preferred trajectory, if the 
planning problem admits solutions. This is consistent and in 
line with the practice used in many well-known planning systems, 
in which only one solution is returned.


\subsection{Computing An Admissible Weight Function}

Let $\Psi$ be a general preference. We will now show how an
admissible weight function $w_\Psi$ can be built in a  
bottom-up fashion. We begin with the basic desires.

\begin{definition}[Basic Desire Weight]
\label{defbasic}
Let $\varphi$ be a basic desire formula and let $\alpha$
be a trajectory. The weight of the trajectory $\alpha$ w.r.t.
the basic  desire $\varphi$ is a function defined as
\[
w_\varphi(\alpha) = \left\{
\begin{array}{lll} 
1 & \textnormal{if } \alpha \models \varphi\\
\\
0 & \textnormal{otherwise}
\end{array}
\right.
\]
\end{definition}
The following proposition derives directly from the
definition of admissibility.
\begin{proposition}\label{basic}
Let $\varphi$ be a basic desire. Then $w_\varphi$ is an admissible weight function.
\end{proposition}
The weight function of an atomic preference builds on the weight function
of the basic desires occurring in the preference as follows.
\begin{definition}[Atomic Preference Weight]
\label{atomicweight}
Let $\psi = \varphi_1 \lhd \varphi_2 \lhd \cdots \lhd \varphi_k$ be
an atomic preference formula. The weight of a trajectory $\alpha$
w.r.t.  $\psi$ is defined as
follows:
\[
   w_{\psi}(\alpha) = 	\sum_{r=1}^{k} ( 2^{k-r} \times w_{\varphi_r}(\alpha) )
\]
\end{definition}
\begin{proposition}
\label{atompref}
Let $\psi = \varphi_1 \lhd \varphi_2 \lhd \cdots \lhd \varphi_k$ be an
atomic preference formula, and let $\alpha_1, \alpha_2$ be two trajectories. Then
\[
\begin{array}{lcr}
\alpha_1 \prec_{\psi} \alpha_2 & \textit{iff} & w_{\psi}(\alpha_1) > w_{\psi}(\alpha_2)
\end{array}
\]
Furthermore, we also have that
\[
\begin{array}{lcr}
\alpha_1 \approx_{\psi} \alpha_2 & \textit{iff} & w_{\psi}(\alpha_1) = w_{\psi}(\alpha_2).
\end{array}
\]
\end{proposition}
\begin{proof}
Let us start by assuming $\alpha_1 \prec_{\psi} \alpha_2$. According to the definition, this
means that $\exists (1 \leq i \leq k)$ such that 
\begin{itemize}
\item $\forall (1 \leq j < i) ( \alpha_1 \approx_{\varphi_j} \alpha_2)$
\item $\alpha_1 \prec_{\varphi_i} \alpha_2$
\end{itemize} 
    From Proposition \ref{basic} we have 
	that $\alpha_1 \approx_{\varphi_j} \alpha_2$ 
	implies $w_{\varphi_j}(\alpha_1) = w_{\varphi_j}(\alpha_2)$
 	for $1 \le j < i$. 
This leads to
$$\sum_{r=1}^{i-1} ( 2^{k-r} \times w_{\varphi_r}(\alpha_1) ) = 
	\sum_{r=1}^{i-1} ( 2^{k-r} \times w_{\varphi_r}(\alpha_2) )$$
In addition, since $\alpha_1 \prec_{\varphi_i} \alpha_2$, then we also have
that $1 = w_{\varphi_i}(\alpha_1) > w_{\varphi_i}(\alpha_2) = 0$. Thus, we have
\[
\begin{array}{lc}
	\sum_{r=1}^{k} ( 2^{k-r} \times w_{\varphi_r}(\alpha_1 ) )  & =\\
	\sum_{r=1}^{i-1} ( 2^{k-r} \times w_{\varphi_r}(\alpha_1) )	 +
		2^{k-i} + \sum_{r=i+1}^{k} ( 2^{k-r} \times w_{\varphi_r}(\alpha_1) ) & > \\
	\sum_{r=1}^{i-1} ( 2^{k-r} \times w_{\varphi_r}(\alpha_1) ) +
	2^{k-i} - 1 & \geq \\
	\sum_{r=1}^{i-1} ( 2^{k-r} \times w_{\varphi_r}(\alpha_1) ) + 
	\sum_{r=i+1}^{k} ( 2^{k-r} \times w_{\varphi_r}(\alpha_2) )  & =\\
	\sum_{r=1}^{k} ( 2^{k-r} \times w_{\varphi_r}(\alpha_2 ) )
\end{array}
\]
For similar reasons, it is also easy to see that 
$\alpha_1 \approx_{\psi} \alpha_2$ implies $w_{\psi}(\alpha_1) = w_{\psi}(\alpha_2)$.

Let us now explore the opposite direction. Let us assume that 
$w_{\psi}(\alpha_1) > w_{\psi}(\alpha_2)$. It is easy to see 
that there must be an integer $i$, $1 \le i \le k$,
such that $\alpha_1 \not\approx_{\varphi_i} \alpha_2$.
Consider the minimal integer $i$ satisfying this property,
i.e., $\alpha_1 \approx_{\varphi_j} \alpha_2$ for every $j$,
$1 \le j < i$. Since  $w_{\psi}(\alpha_1) > w_{\psi}(\alpha_2)$
and $\alpha_1 \not\approx_{\varphi_i} \alpha_2$
we conclude that $\alpha_1 \prec_{\varphi_j} \alpha_2$.
This implies that 
$\alpha_1 \prec_{\psi} \alpha_2$. 

Similar arguments allow 
us to prove that if 
$w_{\psi}(\alpha_1) = w_{\psi}(\alpha_2)$ 
then $\alpha_1 \approx_{\psi} \alpha_2$.
\end{proof}
We are now ready to define an admissible weight function w.r.t. a general preference.
\begin{definition}[General Preference Weight]
Let $\Psi$ be a general preference formula. The weight of 
a trajectory $\alpha$ w.r.t. $\Psi$ 
(denoted by $w_{\Psi}(\alpha)$) is defined as follows:
\begin{list}{$-$}{\topsep=1pt \parsep=0pt \itemsep=1pt}
\item if $\Psi$ is an atomic preference then the weight
	is defined as in Definition \ref{atomicweight}.
\item if $\Psi = \Psi_1 \& \Psi_2$ then 
$	w_{\Psi}(\alpha) = w_{\Psi_1}(\alpha) + 
				w_{\Psi_2}(\alpha)
$
\item if $\Psi = \Psi_1 \mid \Psi_2$ then 
$
	w_{\Psi}(\alpha) = w_{\Psi_1}(\alpha) +
				w_{\Psi_2}(\alpha)
$
\item if $\Psi = \:\:\:!\: \Psi_1$ then 
$
	w_{\Psi}(\alpha) = max(\Psi_1) - w_{\Psi_1}(\alpha)
$
where $max(\Psi_1)$ represents the maximum weight that a
	trajectory can achieve on the preference formulae
	$\Psi_1$ plus one.
\item if $\Psi = \Psi_1 \lhd \Psi_2$\footnote{Because of
Proposition \ref{associative}, without loss of 
generality, we describe the encoding only for chains of length 2.} then
$
   w_{\Psi}(\alpha) = 	 max(\Psi_2) \times w_{\Psi_1}(\alpha) + w_{\Psi_2}(\alpha)
$
\end{list}
\end{definition}
We prove the admissibility of $w_\Psi$ in the next proposition.
\begin{proposition}
\label{p5}
For a general preference $\Psi$, $w_\Psi$ is an 
admissible weight function. 
\end{proposition}
The proof of this proposition is based on Propositions 
\ref{basic}-\ref{atompref} and the structure of $\Psi$. It is omitted here 
for brevity.
The above propositions allow us to prove the following result.
\begin{proposition}
Let $\Psi$ be a general preference and $\alpha$ be 
a trajectory with the maximal weight w.r.t.
$w_{\Psi}$. Then, 
$\alpha$ is a most preferred trajectory w.r.t. $\Psi$.
\end{proposition}
\begin{proof}
Let $w_{\Psi}(\alpha)$ be maximal; let us assume by contradiction
that there exists $\beta$ such that $\beta \prec_{\Psi} \alpha$. It 
follows from the previous proposition that 
$\beta \prec_{\Psi} \alpha$ implies that 
$w_{\Psi}(\beta) < w_{\Psi}(\alpha)$, which contradicts the
hypothesis that $w_{\Psi}(\alpha)$ is maximal.
\end{proof}

Propositions  \ref{basic}-\ref{p5} show that we can compute an admissible 
weight function $w_\Psi$ bottom-up from the weight of each basic desire
occurring in $\Psi$. We are now ready to define the set of rules 
$\Pi_{pref}(\Psi)$, which consists of the rules encoding $\Psi$
and the rules encoding the computation of $w_\Psi$. 
Similarly to the encoding of the
basic desires, we will assign a new, distinguished 
name $n_\phi$ to each preference formula $\phi$
occurring in $\Psi$ and encode the 
preferences in the same way as 
we did for the basic desires (Section \ref{enconde-desire}).
We will also add an atom $preference(n_\phi)$ to the 
set of atoms encoding $\phi$. 
For brevity, we omit here the details of this step. The program  
$\Pi_{pref}(\Psi)$ defines two predicates, $w(p,n)$ and 
$max(p,n)$, where $p$ is a preference name and $n$ is the weight 
of the current trajectory with respect to the preference named $p$. 
$w(p,n)$ (resp. $max(p,n)$) is true if the weight (resp. maximal weight)
of the current trajectory with respect to the preference $p$ is $n$.
\begin{enumerate}
\item For each basic desire $\phi$, $\Pi_{pref}(\phi)$ contains the rules 
encoding $\phi$ and the following rules:
\begin{equation} \label{cw-basic}
\begin{array}{rcl}
w(n_\phi, 1)  & \leftarrow &   satisfy(n_\phi, 0)   \\
w(n_\phi, 0)  & \leftarrow  & \naf satisfy(n_\phi, 0) \\
max(n_\phi, 2) &  \leftarrow &  \label{cw3}
\end{array}
\end{equation}
\item For each atomic preference 
$\phi = \varphi_1 \lhd \varphi_2 \lhd \cdots \lhd \varphi_k$, 
$\Pi_{pref}(\phi)$ consists of 
$$\cup_{j=1}^k \Pi_{pref}(\varphi_k)$$
and the following rules:
\begin{equation} \label{cw-atomic}
\begin{array}{rcl}
w(n_\phi, S) & \leftarrow  & w(n_{\varphi_1}, S_1),  
     \dots,
     w(n_{\varphi_k}, S_k), S = \Sigma^k_{r=1} 2^{k-r} \times S_r
\\
max(n_\phi, 2^{k})  & \leftarrow &
\end{array}
\end{equation}

\item For each general preference $\Psi$, 
\begin{list}{$\bullet$}{\topsep=1pt \parsep=0pt \itemsep=1pt \leftmargin=4pt}
\item if $\Psi$ is an atomic preference then $\Pi_{pref}(\Psi)$ 
is defined as in the previous item.
\item if $\Psi = \Psi_1 \& \Psi_2$ 
or $\Psi = \Psi_1 | \Psi_2$
then $\Pi_{pref}(\Psi)$  consists of 
$$\Pi_{pref}(\Psi_1)\: {\cup}\: \Pi_{pref}(\Psi_2)$$  and 

\begin{equation} \label{cw-and}
\begin{array}{rcl}
w(n_\Psi, S) & \leftarrow &  w(n_{\Psi_1}, N_1),  
     w(n_{\Psi_2}, N_2), S =  N_1 + N_2\\
max(n_\Psi, S)  & \leftarrow &  max(n_{\Psi_1}, N_1),  
     max(n_{\Psi_2}, N_2), S =  N_1 + N_2
\end{array}
\end{equation}
\item if $\Psi = !\: \Psi_1$ 
then $\Pi_{pref}(\Psi)$  consists of 
$\Pi_{pref}(\Psi_1)$  and the rules  
\begin{equation} \label{cw-negation}
\begin{array}{rcl}
w(n_\Psi, S)  & \leftarrow &  w(n_{\Psi_1}, N), max(n_{\Psi_1}, M),
      S =  M - N \\
max(n_\Psi, S)  & \leftarrow &  max(n_{\Psi_1}, S). 
\end{array}
\end{equation}
\item if $\Psi = \Psi_1 \lhd \Psi_2$ 
then $\Pi_{pref}(\Psi)$  consists of 
$\Pi_{pref}(\Psi_1) \cup \Pi_{pref}(\Psi_2)$  and rules 
\begin{equation} \label{cw-chain}
\begin{array}{rcl}
w(n_\Psi, S) & \leftarrow &  w(n_{\Psi_1}, N_1), w(n_{\Psi_2}, N_2), \\
	& &      max(n_{\Psi_2}, M_2),       S =  M_2*N_1 + N_2 \\
max(n_\Psi, S) & \leftarrow &  max(n_{\Psi_1}, N_1),  
     max(n_{\Psi_2}, N_2), \\
	& & S =  N_2*N_1 + N_2 
\end{array}
\end{equation}
\end{list}
\end{enumerate}
The next theorem proves the correctness of $\Pi_{pref}(\Psi)$. 
\begin{theorem} \label{weight-preference-compute}
Let $\langle D,I,G \rangle$ be a planning problem, 
$\Psi$ be a general preference, 
and $\alpha=s_0a_1\ldots a_{n}s_n$ be a trajectory.
Let $\Pi = \Pi_{pref}(\Psi) \cup \Pi_{sat} \cup \alpha^{-1}$.  
Then, 
\begin{itemize}
\item 
For every desire $\phi$ with $desire(n_\phi) \in \Pi_{pref}(\Psi)$, 
we have that 
$\Pi \models w(n_\phi,w)$ iff $w_\phi(\alpha) = w$ and 
$\Pi \models max(n_\phi,w)$ iff $max(\phi)= w$.
\item 
For every preference $\eta$ with $preference(n_\eta) \in \Pi_{pref}(\Psi)$, 
we have that 
$\Pi \models w(n_\eta,w)$ iff $w_\eta(\alpha) = w$
and 
$\Pi \models max(n_\eta,w)$ iff $max(\eta)=w$
\end{itemize}
where 
\[\alpha^{-1} = \{occ(a_i,i-1) \mid i \in \{1,\ldots,n\}\} \cup
\{holds(f,i) \mid f \in s_i, i \in \{0,\ldots,n\}\}.\]
\end{theorem}
\begin{proof}
Let $\Pi_1$ be the program consisting of the rules (\ref{cw-basic})-(\ref{cw-chain}) 
of the program $\Pi_{pref}({\Psi})$ and 
the set of atoms of the form $preference(n_\phi)$ in $\Pi_{pref}({\Psi})$.
Let $S$ be the set of literals occurring in the program $\Pi \setminus \Pi_1$.
It is easy to check that $S$ is a splitting 
set of $\Pi$. Using the Splitting Theorem \cite{lif94a}, we can 
show that $M$ is an answer set of $\Pi$ iff $M = X \cup Y$, where 
$X$ is an answer set of the program $\Pi \setminus \Pi_1$ 
and $Y$ is an answer set of the program $\Pi_2$, which is obtained 
from $\Pi_1$ by replacing the rules of the form 
(\ref{cw-basic}) with the set of atoms $Z$ where
\begin{equation} \label{eq-def-z}
\begin{array}{lll}
Z = \bigcup_{desire(n_\phi) \in X} & 
\{w(n_\phi, 1) \mid satisfy(n_\phi, 0) \in X\}
& \hspace{.5cm}\cup \\
& \{w(n_\phi, 0) \mid satisfy(n_\phi, 0) \not\in X\}
& \hspace{.5cm}\cup \\
& \{max(n_\phi, 2)\}.&
\end{array}
\end{equation}
Observe that for a desire $\phi$ with $desire(n_\phi) \in \Pi_{pref}(\Psi)$
we have that $\Pi_\phi \subseteq \Pi_{pref}(\Psi)$. By 
applying the results of 
Theorem \ref{th-basic-desire}, we have that $\Pi \setminus \Pi_1$ 
has a unique answer set $X$ and $satisfy(n_\phi, 0) \in X$ 
iff $\alpha \models \phi$. Together with the fact that $Z \subseteq Y$,
we have that $w(n_\phi, 1) \in M$ iff $\alpha \models \phi$ 
iff $w_\phi(\alpha) = 1$ and 
$w(n_\phi, 0) \in M$ iff $\alpha \not\models \phi$ 
iff $w_\phi(\alpha) = 0$.
Furthermore, $max(n_\phi,2) \in M$ and 
$w_\phi(\alpha) \le 1$ for every desire $\phi$. This proves the 
first item of the theorem.

We will now prove the second item of the theorem. 
To account for the structure of the preference, we 
associate an integer, denoted by
$\lambda(n_\phi)$,
to each constant $n_\phi$ such that 
$preference(n_\phi) \in \Pi_{pref}(\Psi)$ or 
$desire(n_\phi) \in \Pi_{pref}(\Psi)$. This is done as follows:
\begin{itemize}
\item $\lambda(n_\phi) = 0$ if $desire(n_\phi) \in \Pi_{pref}(\Psi)$
(i.e., if $\phi$ is a desire);
\item $\lambda(n_\phi) = 1$ if $preference(n_\phi) \in \Pi_{pref}(\Psi)$
and $\phi = \varphi_1 \lhd \varphi_2 \lhd \ldots \lhd \varphi_k$; 
\item $\lambda(n_\phi) = \lambda(n_{\phi_1}) + \lambda(n_{\phi_2}) + 1$ 
if $preference(n_\phi) \in \Pi_{pref}(\Psi)$
and $\phi = \phi_1 \& \phi_2$ or
$\phi = \phi_1 \mid \phi_1$; and
\item $\lambda(n_\phi) = \lambda(n_{\phi_1}) + 1$ 
if $preference(n_\phi) \in \Pi_{pref}(\Psi)$
and $\phi = ! \phi_1$.
\end{itemize}
The proof is done inductively over $\lambda(n_\phi)$. 
\begin{itemize}
\item {\bf Base:} $\lambda(n_\phi) = 0$ means that $n_\phi$ is a desire.
The claim for this case follows from the first item. 
\item {\bf Step:} Assume that we have proved the conclusion for 
$\lambda(n_\phi) < k$. We will now prove it for $\lambda(n_\phi) = k$.
Consider a preference $\phi$ with $\lambda(n_\phi) = k$. We have 
the following cases:
\begin{itemize}
\item $\phi = \varphi_1 \lhd \varphi_2 \lhd \ldots \lhd \varphi_k$
and $\varphi_i$ are basic desires, i.e., $\phi$ is an atomic 
preference. 
By definition, we have that $\varphi_i$'s are desires. 
It follows from (\ref{cw-atomic}) that \\
\centerline{$w(n_\phi, s) \in Y$ iff 
the body of the first rule in (\ref{cw-atomic}) is satisfied by $Y$} \\
\centerline{iff $s = \Sigma_{r=1}^l 2^{k-r} \times w_r$ 
and $w(n_{\varphi_i}, w_r) \in Z$ for $1 \le i \le k$} \\
\centerline{iff $s = w_\phi(\alpha)$.}

\noindent
Furthermore, $max(n_\phi, 2^k) \in Y$ and 
the maximal weight of $w_\phi(\alpha)$ is 
$\Sigma_{r=1}^k 2^{k-r} = 2^k -1$. This proves 
the inductive step for this case.

\item $\phi = \phi_1 \& \phi_2$ (resp. $\phi = \phi_1 \mid \phi_2$). 
We have that $\lambda(n_{\phi_1}) < k$ and $\lambda(n_{\phi_2}) < k$.
The conclusion follows immediately from the induction hypothesis, 
the rules in (\ref{cw-and}),
and the definition of $w_\phi(\alpha)$. 

\item $\phi = ! \phi_1$. Again,  we have that $\lambda(n_{\phi_1}) < k$. 
Using (\ref{cw-negation}) and the definition of 
$w_\phi$, we can prove that $w(n_\phi, s) \in Y$ 
iff $w_\phi(\alpha) = s$ and $max(n_\phi, s) \in Y$ 
iff $max(\phi) = s$

\item $\phi = \phi_1 \lhd \phi_2$. Again, 
we have that $\lambda(n_{\phi_1}) < k$ and $\lambda(n_{\phi_2}) < k$.
The conclusion follows immediately from (\ref{cw-chain}),
the induction hypothesis, and the definition of $w_\phi(\alpha)$. 
\end{itemize}
\end{itemize}
\end{proof}
The above theorem implies that we can compute a most preferred trajectory 
by (i) adding $\Pi_{pref}(\Psi) \cup \Pi_{sat}$ to $\Pi(D,I,G)$ and
(ii) computing an answer set $M$ in which $w(n_\Psi, w)$ is maximal. 
A working 
implementation of this is available in {\bf jsmodels}.

\subsection{Some Examples of Preferences in $\cal PP$}

We will now present some preferences that are common to many planning 
problems and have been discussed in \cite{eiter02a}. The main 
difference between the encoding presented in this paper and the 
ones in \cite{eiter02a} lies in that we use temporal operators to 
represent the preferences, while action weights are used in 
\cite{eiter02a}. 
Let $\langle D,I,G \rangle$ be a planning problem. For the 
discussion in this subsection, we will assume that the answer 
set planning module $\Pi(D,I,G)$ is capable of generating trajectories
without redundant actions in the sense that no action occurrence is generated 
once the goal has been achieved. Such a planning module can be easily 
obtained by adding a constraint to the program $\Pi(D,I,G)$ 
preventing action occurrences to be generated once the goal 
has been achieved. This, however, does not guarantee that the
planning module will generate the shortest trajectory if $n$ is 
greater than the length of the shortest trajectory. 
In keeping with the notation used 
in the previous section, we use $\varphi$ to denote $G$ (i.e., 
$\varphi = G$).  

\subsubsection{Preference for shortest trajectory --  formula based encoding} 

Assume that we are interested in trajectories achieving $\varphi$ 
whose length is less than or equal $n$. 
A simple encoding that allows us to accomplish such goal
is to make use of basic desires. By 
${\bf next}^i(\varphi)$ we denote the formula:
$$
 \underbrace{{\bf next} ( {\bf next} ( {\bf next} \cdots ( {\bf next}}_{i} (\varphi)) \cdots)). $$
Let us define the formula
$\sigma^i(\varphi)$ ($0 \leq i \leq n$) as follows:
$$
\begin{array}{lcl}
\sigma^0(\varphi)   =  \varphi & \hspace{2cm} &
\sigma^i(\varphi)   =  \bigwedge_{j=0}^{i-1} \neg {\bf next}^j(\varphi) \:\: \wedge \:\:
				{\bf next}^i(\varphi)
\end{array}
$$
Finally, let us consider the formula $short(n,\varphi)$ defined as
$$
	short(n,\varphi) = \sigma^0(\varphi) \lhd \sigma^1(\varphi) 
		\lhd \sigma^2(\varphi) \lhd \cdots \lhd \sigma^n(\varphi).
$$
Intuitively, this formula says that we prefer trajectories on which 
the goal $\varphi$ is satisfied as early as possible. 
It is easy to see that if $\alpha$ is a most preferred trajectory 
w.r.t. $short(n,\varphi)$ then 
$\alpha$ is a shortest length trajectory satisfying the goal $\varphi$.

\subsubsection{Preference for shortest trajectory -- action based encoding} 
The formula based encoding $short(n,\varphi)$ requires the bound 
$n$ to be given. We now present another encoding that does not
require this condition. We introduce two additional fictitious 
actions $stop$ and $noop$ and a new fluent $ended$. The action
$stop$ will be triggered when the goal is achieved; $noop$ is used
to fill the slot so that we can compare between trajectories;
the fluent $ended$ will denote the fact that the goal has 
been achieved.
We add to the action theory the 
propositions: 
\[ 
\begin{array}{l}
stop \:\: {\bf causes} \:\: ended \\
stop \:\: {\bf executable \ if} \:\: \varphi\\
noop \:\: {\bf causes} \:\: ended\\
noop \:\: {\bf executable \ if} \:\: ended
\end{array}
\]
Furthermore, we add the condition $\neg ended$ to the
executability condition of every action in $(D,I)$ 
and to the initial state $I$.
We can encode the condition of shortest length trajectory
as follows. Let 
$$
short = {\bf always}((stop \vee noop) <^e (a_1 \vee \dots \vee a_k)).
$$
where $a_1, \dots, a_k$ are the actions in the original action theory.
Again, we can show that any most preferred 
trajectory w.r.t. $short$ 
is a shortest length trajectory satisfying the goal $\varphi$.
Observe the difference between $short(n,\varphi)$ and $short$:
both are built using temporal connectives but the former 
uses fluent 
formula and the latter uses actions. The second one, we believe,
is simpler than the first one; however, it requires some 
modifications to the original action theory.

\subsubsection{Cheapest plan}
Let us assume that we would like to associate a cost $c(a)$ to 
each action $a$  and determine trajectories that have the minimal cost.
Since our comparison is done only on trajectories whose length is 
less than or equal {\em length},
we will also introduce the two actions $noop$ and $stop$ with no 
cost and the fluent $ended$ to record the fact that the goal 
has been achieved. Furthermore, we introduce the fluent $sCost(ct)$
to denote the cost of the trajectory. Intuitively,
$scost(ct)$ is true mean that the cost of the trajectory is $ct$.
Initially, we set the value of $sCost$ to $0$
(i.e., $sCost(0)$ is true initially and 
$sCost(c)$ is false for every other $c$) and the execution of action $a$ will 
increase the value of $sCost$ by $c(a)$. This is done by 
introducing an effect proposition
\[
a \causes sCost(N+c(a)) \iif sCost(N)
\]
for each action $a$\footnote{
   Because of the grounding requirement of answer set solver,
   this encoding will yield a set of effect propositions 
   instead of a single proposition. 
}.
The preference
$$\goal(sCost(m)) \lhd \goal(sCost(m+1)) 
\ldots \lhd \goal(sCost(M))
$$
where $m$ and $M$ are the estimated minimal and maximal 
cost of the trajectories, respectively. Note that we can have 
$m = 0$ and $M = max\{c(a) \mid a$ is an action$\} \times length$.

\section{Related Work} \label{related}
The work presented in this paper is the
natural continuation of the work we presented in  
\cite{son-prio}, where we rely on  prioritized default theories
to express limited classes of preferences between trajectories---a
strict subset of the preferences covered in this paper.
This work is also influenced by other works on exploiting 
\emph{domain-specific knowledge}
in planning (e.g., \cite{bacchus00,lago,sbm02a}),
in which domain-specific knowledge is expressed as a constraint 
on the trajectories achieving the goal, and hence, is a {\em hard
constraint}. In subsection \ref{planpref}, we discuss 
different approaches to planning with preferences which are 
directly related to our work. In
Subsections \ref{highlevel}--\ref{other} we present
works that are somewhat related to our work and can be used 
to develop alternative implementation for $\cal PP$.  

\subsection{Planning with Preferences}
\label{planpref}
Different approaches have been proposed to integrate preferences
in the planning process.
An approach close in spirit to the one proposed in this paper has
been recently developed by Delgrande et al. \shortcite{delgrande-plan}. The
framework they propose introduces qualitative preferences built from 
two partial preorders, $\leq_c$ and $\leq_t$, over the set of propositional 
formulae of fluents and actions. Intuitively,
\begin{itemize}
\item  $\varphi_1 \leq_c \varphi_2$ (\emph{choice order}) indicates
the desire to prefer trajectories that satisfy (at some point in time) the formula
$\varphi_2$ over those that satisfy $\varphi_1$. 
\item $\varphi_1 \leq_t \varphi_2$ (\emph{temporal order}) indicates
the desire to prefer trajectories that satisfy $\varphi_1$ first and 
	$\varphi_2$ later in the trajectory.
\end{itemize}
Choice preferences are employed to derive an ordering $\lhd_c$ between trajectories
as follows: given trajectories $\alpha,\beta$ we have that
\[\alpha \lhd_c \beta \:\:\textit{iff}\:\: \forall \varphi \in \Delta(\alpha,\beta). 
	\exists \varphi' \in \Delta(\beta,\alpha). (\varphi \leq_c \varphi')\]
where $\Delta(\gamma,\gamma') = \{\varphi \in dom(\leq_c)\:|\: \gamma\models \varphi, \gamma' \not\models \varphi\}$
and $\gamma \models \varphi$ denotes the fact that the formula $\varphi$ is true at
one of the states reached by the trajectory $\gamma$.
The order is made transitive by taking the transitive closure of $\lhd_c$.

The relation $\lhd_c$ can be easily simulated in $\cal PP$ since
 $\varphi \leq_c \varphi'$
determines the same order as 
$\eventually(\varphi') \lhd \eventually(\varphi)$. This can be generalized
as long as $\leq_c$ is cycle-free.

\begin{example}\label{ex6}
Let us consider the monkey-and-banana example as formulated in 
\cite{delgrande-plan}.
The world includes the following entities: a monkey, a banana hanging from the ceiling, a 
coconut on the floor, and a chocolate bar inside a closed drawer. Initially, all the entities are in different locations
in a room. The room includes also a box that can be pushed and climbed on to reach the ceiling and grab
the banana. The goal is to get the chocolate as well as at least one of 
the banana or the coconut. The domain description includes the following fluents:
\begin{itemize}
\item $location(Entity, Location)$ denoting the current $Location$ of $Entity$; the 
	domain of $Entity$ is $\{monkey, banana, coconut, drawer, box\}$ and the
	domain of $Location$ is $\{1,\dots,5\}$ (denoting $5$ different positions in the
	room).
\item $onBox$ denoting the fact that the monkey is on top of the box.
\item $hasBanana$ denoting the fact that the monkey has the banana.
\item $hasCoconut$ denoting the fact that the monkey has the coconut.
\item $hasChocolate$ denoting the fact that the monkey has the chocolate.
\item $DrawerOpen$ denoting the fact that the drawer is open.
\end{itemize}

The action theory provides actions to walk in the room, move the
box, climb on and off the box, grab objects, and open drawers.
The goal considered here is expressed by the fluent formula:
\[ hasChocolate \wedge (hasCoconut \vee hasBanana) \]

The preference discussed in \cite{delgrande-plan} is that bananas are preferred over
coconuts---i.e., $hasCoconut \leq_c hasBanana$---and in our framework it can be expressed as 
\[ \eventually(hasBanana) \lhd \eventually(hasCoconut). \]
This preference can also be represented by a simpler basic desire
\[ \goal(hasBanana) \]
which says that trajectories achieving $hasBanana$ will be most preferred. 
\hfill$\Box$
\end{example}
Temporal preferences are employed to derive another preorder $\lhd_t$ between
trajectories as follows:
\begin{itemize}
\item given a trajectory $\alpha = s_0 a_1 \ldots a_n s_n$ 
	and two propositions over fluent and actions $\varphi, \varphi'$,
	then $\varphi \leq_{\alpha} \varphi'$ iff 
	\begin{itemize}
	\item $\alpha \models \varphi$ and $\alpha \models \varphi'$ and
	\item $i_{\varphi} \leq i_{\varphi'}$ where 
		$s_{i_{\varphi}}$ (resp. $s_{i_{\varphi'}}$) 
		is the first state  in $\alpha$
		that satisfies $\varphi$ (resp. $\varphi'$). 
	\end{itemize}
\item given two trajectories $\alpha, \beta$, we have that $\alpha \lhd_t \beta$ iff
		$<_t \cap \leq_{\beta}^{-1} \subseteq <_t \cap \leq_{\alpha}^{-1}$
	where $\leq_{\alpha}^{-1}$ is the inverse relation of $\leq_\alpha$.
\end{itemize}
Each individual temporal preference $\varphi \leq_t \varphi'$ can be expressed
in our language as the basic desire
\[ c(\varphi\leq_t \varphi') \equiv \eventually(\varphi \wedge \eventually(\varphi')) \wedge
	\until(\neg \varphi',\varphi)\]
The generalization to a collection of temporal preferences requires some additional
constructions. Given a collection of basic desires $S=\{\psi_1,\dots,\psi_k\}$, then
\begin{itemize}
\item for an arbitrary permutation $i_1,\dots,i_k$ of $1,\dots,k$, let us
	define
\[ch(S, i_1,\dots,i_k) \equiv 
	\bigwedge_{j=1}^k \psi_{i_j} \lhd 
	\bigwedge_{j=2}^k \psi_{i_j} \lhd
	\bigwedge_{j=3}^k \psi_{i_j} \lhd \cdots \lhd
	\psi_{i_k}.
\]
Intuitively, $ch(S, i_1,\ldots,i_k)$ is an atomic preference 
representing an ordering between trajectories w.r.t. the set of
basic desires $\{p_{i_1}, \dots, p_{i_k}\}$. For example, 
trajectories  satisfying $\bigwedge_{j=1}^k \psi_{i_j}$ is most 
preferred; if no trajectory satisfies $\bigwedge_{j=1}^k \psi_{i_j}$
then trajectories satisfying $\bigwedge_{j=2}^k \psi_{i_j}$ is most 
preferred; etc.

\item let $\{\pi_1,\dots,\pi_{k!}\}$ be 
the set of all permutations of $1,\dots,k$; let us
	define
\[ maxim(S) \equiv ch(S, \pi_1) \:|\: ch(S, \pi_2) \:|\: \cdots \:|\: ch(S, \pi_{k!}).\]
Intuitively, $maxim(S)$ indicates that we prefer 
trajectories satisfying the maximal number of 
basic desires from the set $\{\psi_1,\ldots,\psi_k\}$. 
\end{itemize}
If we have a collection of temporal preferences 
$\{\varphi_i \leq_t \varphi_i'\:|\: i=1,\dots,k\}$, then the equivalent formula
is
\[ maxim(\{ c(\varphi_i \leq_t \varphi_i') \:|\: i=1,\dots,k\}).\]

\begin{example}
Let us continue Example \ref{ex6} by removing the choice preference and
assuming instead the temporal preference $hasBanana \leq_t hasChocolate$---i.e.,
the banana should be obtained before the chocolate. The corresponding encoding in
our language is
\[\begin{array}{rl}
	 \eventually(hasBanana \wedge \eventually(hasChocolate)) & \wedge\\ 
	\multicolumn{2}{r}{\until(\neg hasChocolate, hasBanana)}
  \end{array}
\]
\hfill$\Box$
\end{example}

\smallskip
Eiter et al. introduced a framework for planning
with action costs
using logic programming \cite{eiter02a}. The focus of their
proposal is to express certain classes of quantitative preferences.
Each action is assigned an integer cost, and 
plans with the minimal cost are considered to be optimal. 
Costs can be either static or relative to the 
time step in which the action is executed. \cite{eiter02a}
also presents the encoding of different preferences, 
such as shortest plan and the cheapest plan. Our approach 
also emphasizes the use of logic programming,  but differs in
several aspects.
Here, we develop a \emph{declarative language} for preference representation.
Our language can express 
the preferences discussed in \cite{eiter02a}, but
 it is more high-level
 and flexible than the action costs approach. 
The approach in \cite{eiter02a} also
does not allow the use of  fully general 
dynamic preferences. On the other hand, while we
only consider planning with 
complete information,  Eiter et al. \cite{eiter02a} 
deal with planning in the presence 
of incomplete information and non-deterministic actions. 

Other
systems have adopted fixed types of preferences, e.g., 
shortest plans \cite{cimatti,blumfurst}.

Our proposal has  similarities with the 
approach based on metatheories of the planning
domain \cite{myers1,myers99a}, where metatheories provide
characterization of semantic differences between the various
domain operators and planning variables; metatheories
allow the generation of biases to focus the planner 
towards plans with certain characteristics.

Our work is also  related to the work in \cite{lin98}
in which the author defined three different measures for 
plan quality (A-, B-, and C-optimal) and showed how they 
can be axiomatized in situation calculus. Roughly, a 
plan is A-optimal if none of its actions can be deleted
and the remainder is still a valid plan. It is B-optimal if 
none of its segments can be deleted and the remainder is still 
a plan. It is C-optimal if none of its segments can be replaced 
by a single action and the remainder is still a plan. 
While these measures are domain-independent, preferences in our 
language are mostly domain-dependent. Theoretically, these measures 
could also be expressed in $\cal PP$ by defining an order among 
possible plans. This impractical method can be 
replaced by considering some approximations of these measures. 
As an example, 
shortest plans as encoded in the previous section represents a class
of A-optimal plans; the atomic preference $\varphi_1 \lhd \varphi_2$
with $\varphi_1 = occ(a) \wedge \mathbf{executable}(a)$ 
and $\varphi_2 = occ(b)  \wedge \mathbf{executable}(b) 
		\wedge \next(occ(c)  \wedge \mathbf{executable}(c))$  
could be used to prefer plans with action $a$ over plan containing 
the sequence $b;c$; etc.

Our language allows the representation of several types of 
preferences similar to those developed 
in \cite{haddawy93a} for decision-theoretic planners. 
The fundamental difference is that we use logic programming while their 
system is probability based.
Our approach  also differs from the works on using Markov
Decision Processes (MDP) to find optimal plans \cite{puterman94};
in MDPs, optimal plans are functions from states to actions, thus
preventing the user from selecting preferred trajectories without
changing the MDP specification.

\subsection{High-level Languages for Qualitative Preferences}
\label{highlevel}

Brewka recently proposed \cite{brewka-qualitative} a general rank-based
description language for the representation of qualitative preferences between
models of a propositional theory. The language has similar foundations to 
our proposal. The basic preference between models derives from an inherent
total preorder between propositional formulae (\emph{Ranked Knowledge Base}); models
can be compared according to one of four possible comparison criteria---i.e.,
\emph{inclusion}, \emph{cardinality}, \emph{maximal degree of satisfied formula},
and \emph{maximal degree of unsatisfied formula}.
The preference language allows the refinement of the basic preference by using
propositional combination as well as meta-ordering between preferences, in a fashion
similar to what described in this paper.

The proposal by Junker \cite{junker-config} presents a 
language designed to express preferences between decisions and
decision rules in the context of a language for solving configuration problems.
Decisions are described by labeled constraints $t \: : \:\varphi$, where
$t$ is a term (possibly containing variables) and $\varphi$ is a configuration
constraint. The configuration language allows also the creation of named
sets of decisions.
Preferences between decisions are expressed through statements of the
form $prefer(t_1,t_2)$, where $t_1, t_2$ identify decisions or sets of 
decisions. The language allows the user also to create constraints that
assert decisions, thus making it possible to express meta-preferences. For 
example, the following decisions express different preferences \cite{junker-config}:

\begin{verbatim}
    decision rule p1(x):
       if x in instances(Customer) and playboy in characteristics(x) 
            then prefer(look, comfort)
    decision rule p2(x):
       if x in instances(Customer) and age(x)=old 
            then prefer(comfort,look)
\end{verbatim}

\noindent
and a statement of the type $prefer(p1,p2)$ allows to express 
a meta-preference.

\subsection{Other Related Works}
\label{other}

Considerable effort has been invested in developing frameworks for 
expressing preferences within the context of constraint programming and
constraint logic programming, where the problem of inconsistency arises
frequently. Most proposals rely on the idea of associating preferences
(expressed as mathematical entities) to constraints when variables are
assigned \cite{schiex-survey}. Combinations of constraints lead to 
corresponding combinations of preferences, and the frameworks provide
means to compare preferences; comparisons are commonly employed to 
select solutions or to discriminate between classes of satisfied constraints.
A popular scheme relies on the use of \emph{costs} associated to tuples (where
each tuple represent a value assignment), and costs are drawn from a 
semiring structure \cite{schiex1,bista-soft}, which provides operators to 
combine preferences and to ``maximize'' preferences. These frameworks subsume
various approaches to preferences in CSP, e.g.,
\cite{borning-hierarchy,moulin88,fargier}.

Schiex et al. \cite{schiex2,schiex-survey} recently proposed the notion of 
\emph{Valued CSP} as an algebraic framework for preferences in 
constraint network.
In VCSP, costs for tuples are drawn from a value structure 
$\langle E, \oplus, \succeq \rangle$, where $E$ is totally ordered by $\succeq$; 
the maximum denotes total inconsistency. Intuitively, in a valued CSP, each constraint
$c_X$ over a set of variables $X$ is viewed as a function that maps tuples
of values (values drawn from the domains of the variables $X$) to an element of $E$
(the ``cost'' of the tuple). The cost of the constraint allows us to rank the
``degree'' of constraint violation. Given an assignment  $t$ for a set of constraints
$C$, the valuation of the assignment is the $\oplus$ composition of the $E$ values
of the individual constraints. The objective is to determine an assignment which
is minimal w.r.t. the order $\succeq$. \emph{Weighted CSP (WCSP)} are instances
of this framework, where $E = [ 0, 1, \dots, k ]$, $\succeq$ is the standard
ordering between natural numbers, and $\oplus$ is defined as 
$a\oplus b = min\{k, a+b\}$. Extensions of arc-consistency to these frameworks
have been investigated \cite{larrosa,schiex2,bista-soft}.

\medskip

Qualitative measures of preference in constraint programming have been explored
through the notion of \emph{Ceteris Paribus Networks (CP-nets)} \cite{cpnet}.

A CP-net is a graphical tool to represent qualitative preferences. Let $\cal V$
be a set of variables and let us denote with $D(v)$ the domain of variable
$v$ ($v \in {\cal V}$). A CP-net is a pair $\langle G,P \rangle$, where $G$ is
a directed (typically acyclic) graph whose vertices are elements of $\cal V$,
while the edges denote preferential dependences between variables; intuitively,
preferences for a value for a variable $v$ depend only on the values selected
for the parents of $v$ in the network. For a given assignment of values to the
parents of $v$, the CP-net specifies a total order on $D(v)$.
An assignment of values $\gamma$ to
$\cal V$ is immediately preferred to the assignment $\eta$ if there is 
a variable $v$ such that 
\begin{itemize}
	\item $\forall u \in {\cal V}\setminus \{v\}. \: \gamma(u) = \eta(u)$
	\item $\gamma(v)$ is preferred to $\eta(v)$ in the ordering 
	of $D(v)$ specified by the assignment $\gamma$ to the parents of $v$.
\end{itemize}
In general, an assignment $\gamma$ is CP-preferred to an assignment $\eta$ if
there exists a sequence of assignments $\gamma_0, \gamma_1, \dots, \gamma_k$ such
that
\begin{itemize}
	\item $\gamma_0 = \eta$
	\item $\gamma_i$ is immediately preferred to $\gamma_{i-1}$
	\item $\gamma_k = \gamma$
\end{itemize}
Algorithms for solving constraint optimization problems under preference
ordering specified by CP-nets have been proposed in the literature
\cite{cpnet1,cpnet2}.

Constraint solving has also been
proposed  for the management of planning in
presence of action costs \cite{kautz1}.

\bigskip
Considerable effort has been invested in introducing
	preferences in logic programming. In \cite{cui} 
	preferences are expressed at the level of atoms and
	used for parsing disambiguation in logic grammars. Rule-level
	preferences have been used in various proposals for 
	selection of preferred answer sets in answer set 
	programming \cite{brew99,delschtomp03,gl98,schaub}. 
Some of the existing answer set solvers include limited forms of
(numerical) optimization capabilities. {\bf smodels} \cite{smodels-ai} offers
the ability to associate \emph{weights} to atoms and to compute answer
sets that minimize or maximize the total weight. DLV \cite{dlv-constraints}
provides the notion of \emph{weak constraints}, i.e., constraints of the
form 
\[ \leftarrow \ell_1, \dots, \ell_k. \: [ w \: : \: l] \]
where $w$ is a numeric penalty for violating the constraint, and
$l$ is a priority level. The total cost of violating constraints at
each priority level is computed, and answer sets are compared to minimize
total penalty (according to a lexicographic ordering based on priority
levels).

\subsection{Alternative Encodings of ${\cal PP}$}

In this section we discuss the possibility of implementing ${\cal PP}$ using
inference back-ends different from {\bf smodels} or {\bf jsmodels}. It should 
be noted that the encoding proposed in Section \ref{implementation}
can be translated into {\bf dlv} code with little effort,
while it is not so with other answer set programming systems
(e.g., {\bf cmodels}, {\bf ASSAT}),  since they do not offer a construct
similar to the {\bf maximize} construct of {\bf smodels}. 

In this section, we 
explore the relationships between ${\cal PP}$ and two relatively
new answer set programming frameworks, 
\emph{Logic Programming with Ordered Disjunctions} and 
\emph{Answer
Set Optimizations}. The key idea in both cases, is to show
that each preference of ${\cal PP}$ can be mapped to a collection
of rules in these two languages. 
Below we provide some details of these languages and their
use for expressing ${\cal PP}$.

\subsubsection{Logic Programming with Ordered Disjunctions (LPOD)}
\label{lpod}

\paragraph{\underline{Overview of LPOD:}}
In \emph{Logic Programming with Ordered Disjunctions} \cite{lpod}, a program
is a collection of ground rules of the form
\[ A_1 \times \cdots \times A_k \leftarrow B_1, \dots, B_n, \nnot C_1, \dots, \nnot C_m \]
The literals in the head of the rule represent alternative choices; in the specific
case of LPOD, the choices are ordered, where $A_1$ is the most preferred choice, while
$A_k$ is the least preferred one.

The semantics of a LPOD program $P$ is based on the general idea of answer sets and the
concept of \emph{split} of a program. For each rule 
$A_1 \times \cdots \times A_k \leftarrow Body$, the $i^{th}$ option of the rule
is the standard logic programming clause 
\[ A_i \leftarrow Body, \nnot A_1, \dots, \nnot A_{i-1} \]
A split of the program $P$ is a standard logic program obtained by replacing
each rule of $P$ by one of its options.

Given a LPOD program $P$ and a set of ground literals $S$, then $S$ is an answer
set of $P$ iff $S$ is an answer set of a split of $P$.

Ordered disjunctions are employed to create a preference order between answer
sets of a program. Different ordering criteria have been discussed \cite{lpod}.
Given an answer set $S$ of a LPOD program $P$, we say that $S$ satisfies a rule $r$
\[ A_1 \times \cdots \times A_k \leftarrow Body\]
\begin{itemize}
\item with degree 1 ($deg_S(r)=1$) if $S \not\models Body$
\item with degree $i$ ($deg_S(r)=i$) if $S \models Body$ and $i=min\{j \:\mid\: S \models A_j\}$
\end{itemize}
We denote with $S^i(P) = \{r\in P \: | \: deg_S(r) = i\}$.
The three criteria for comparing answer sets under LPOD are the following. Let $S_1,S_2$ be
two answer sets of $P$;
\begin{itemize}
\item $S_1$ is cardinality preferred to $S_2$ ($S_1 >_c S_2$) iff $\exists i$ such
	that $|S_1^j(P)| = |S_2^j(P)|$ for $j < i$ and $|S_1^i(P)| > |S_2^i(P)|$.
\item $S_1$ is inclusion preferred to $S_2$ ($S_1 >_i S_2$) iff $\exists i$ such
	that $S_1^j(P) = S_2^j(P)$ for $j < i$ and $S_1^i(P) \supset S_2^i(P)$.
\item $S_1$ is Pareto preferred to $S_2$ ($S_1 >_p S_2$) iff 
	\begin{itemize}
	\item $\exists r \in P. \: deg_{S_1}(r) < deg_{S_2}(r)$
	\item $\not\exists r' \in P. \: deg_{S_1}(r) > deg_{S_2}(r)$
	\end{itemize}
\end{itemize}

LPOD allows also the use of meta-preferences between rules of the form $r_1 \succ r_2$. The
Pareto preference in this case is modified as follows: $S_1 >_p S_2$ iff
	\begin{itemize}
	\item $\exists r \in P. \: deg_{S_1}(r) < deg_{S_2}(r)$
	\item $\forall r \in P$, if $deg_{S_1}(r) > deg_{S_2}(r)$ then there exists
		another rule $r'$ such that $r' \succ r$ and $deg_{S_1}(r') < deg_{S_2}(r)$
	\end{itemize}

\paragraph{\underline{Translation of our Preferences:}}
Let us start by providing a logic programming encoding of basic desires. We define two
entities: $Core^{\psi}(T)$ is a unary predicate while $rules^{\psi}(T)$ is a
collection of rules where $T$ is a variable representing time step. 
\begin{itemize}
\item if $\psi\equiv occ(a)$ ($a \in \mathbf{A}$) then
	$Core^{\psi}(T) = occ(a,T)$ and
	$rule^{\psi}(T) = \emptyset$.

\item if $\psi\equiv f$ ($f$ is a fluent literal) then 
	$Core^{\psi}(T) = holds(f,T)$ and 
	$rule^{\psi}(T) = \emptyset$.

\item if $\psi\equiv\psi_1 \wedge \psi_2$ then
	$Core^{\psi}(T) = p^{\psi}(T)$ and  
\[ rule^{\psi}(T) = \left\{\begin{array}{l}
			 \leftarrow p^{\psi}(T), \nnot Core^{\psi_1}(T)\\
			\leftarrow p^{\psi}(T), \nnot Core^{\psi_2}(T)\\
			p^{\psi}(T) \leftarrow Core^{\psi_1}(T), Core^{\psi_2}(T)
			   \end{array}
			\right\}
\]
	where $p^{\psi}$ is a new unary predicate.

\item if $\psi\equiv \psi_1 \vee \psi_2$ then
	$Core^{\psi}(T) = p^{\psi}(T)$ and
\[ rule^{\psi}(T) = \left\{ 
			\begin{array}{l}
\leftarrow p^{\psi}(T), \nnot Core^{\psi_1}(T), \nnot Core^{\psi_2}(T)\}\\
	p^{\psi}(T) \leftarrow Core^{\psi_1}(T)\\
	p^{\psi}(T) \leftarrow Core^{\psi_2}(T)
			\end{array}
	\right\}
\]
	where $p^{\psi}$ is a new unary predicate.

\item if $\psi\equiv \neg \psi_1$ then
	$Core^{\psi}(T) = p^{\psi}(T)$ and
\[ rule^{\psi}(T) = \left\{ \begin{array}{l}
	p^{\psi}(T) \leftarrow \nnot Core^{\psi_1}(T)\\
	\leftarrow p^{\psi}(T), Core^{\psi_1}(T)
			    \end{array}
	\right\}
\]
	where $p^{\psi}$ is a new unary predicate.

\item if $\psi \equiv \next(\psi_1)$ then
	$Core^{\psi}(T) = p^{\psi}(T)$ and
\[ rule^{\psi}(T) = \left\{\begin{array}{l}
	 \leftarrow p^{\psi}(T), \nnot Core^{\psi_1}(T+1)\\
	p^{\psi}(T) \leftarrow Core^{\psi_1}(T+1)
			   \end{array}
	\right\}
\]
	where $p^{\psi}$ is a new unary predicate.

\item if $\psi \equiv \always(\psi_1)$ then
	$Core^{\psi}(T)=p^{\psi}(T)$ and
\[ rule^{\psi}(T) = \left\{\begin{array}{l}
		\leftarrow p^{\psi}(T), \nnot Core^{\psi_1}(T1), T \leq T1, T1 \leq n\\
		p^{\psi}(T) \leftarrow always^{\psi_1}(T)\\
		always^{\psi_1}(n) \leftarrow Core^{\psi_1}(n)\\
		always^{\psi_1}(T) \leftarrow T < n, Core^{\psi_1}(T), 
		always^{\psi_1}(T+1) 
			   \end{array}
		\right\}
\]
	where $p^{\psi}$ is a new unary predicate.

\item if $\psi \equiv \eventually(\psi_1)$ then
	$Core^{\psi}(T) = p^{\psi}(T)$ and
\[ rule^{\psi}(T) = \left\{ 
	\begin{array}{l}
		\leftarrow p^{\psi}(T), \nnot Core^{\psi_1}(T), \nnot Core^{\psi_1}(T+1), \dots,
						Core^{\psi_1}(n)\\
		p^{\psi}(T) \leftarrow Core^{\psi_1}(T1), T \leq T1, T1 \leq n
	\end{array}
	\right\}
\]
	where $p^{\psi}$ is a new unary predicate.
\end{itemize}
Let us define as $\Pi^{\psi}(i) = rule^{\psi}(i)\cup\{Core^{\psi}(i) \times \neg Core^{\psi}(i)\}$ and
let us denote with $r^{\psi}(i)$ the rule $Core^{\psi}(i) \times \neg Core^{\psi}(i).$
We can show that if 
$\langle D,I,G\rangle$ is a planning problem and 
$\psi$ is a basic desire, then 
the following holds:
\[ \begin{array}{ccc}
S_1 >_p S_2 & \textit{  iff  } & \pi(S_1) \models \psi \textit{ and } \pi(S_2) \not\models \psi 
   \end{array}
\]
where $S_1,S_2$ are two answer sets of $\Pi(D,I,G)\cup \Pi^{\psi}(i)$
and $\pi(S)$ denotes the trajectory represented by $S$.

Let us extend the encoding above to include atomic preferences. In particular,
given an atomic preference of the form $\psi_1 \lhd \psi_2$, we define
\[ \Pi^{\psi}(i) = rule^{\psi_1}(i) \cup rule^{\psi_2}(i) \cup
		\left\{\begin{array}{ll}
			(r1) & Core^{\psi_1}(i) \times \nnot Core^{\psi_1}(i)\\
			(r2)& Core^{\psi_2}(i) \times \nnot Core^{\psi_2}(i)\\
			r1 \succ r2
		       \end{array}
		\right\}
\]
A result similar to the one above can be
derived: for a planning problem $\langle D,I,G \rangle$ and an atomic preference $\psi$, 
if $S_1,S_2$ are two answer sets of $\Pi(D,I,G) \cup \Pi^{\psi}(i)$, then 
\[ \begin{array}{ccc}
S_1 >_p S_2 & \textit{  iff  } &
	\pi(S_1) \prec_{\psi} \pi(S_2). 
   \end{array}
\]

The encoding of general preferences in the LPOD framework does not appear
to be as simple as in the previous cases. The encoding is clearly possible---it
is sufficient to make use of the encoding presented in Section \ref{implementation}; if
$\psi$ is the preference and $max(n_{\psi},v)$ is true, then we can introduce the rule
\[ w(n_{\psi},v) \times w(n_{\psi},v-1) \times \cdots \times w(n_{\psi},1).\]
The resulting encoding, on the other hand, is not any simpler than the
direct encoding in {\bf smodels} with atom weights.

\subsubsection{Answer Set Optimization (ASO)}

\paragraph{\underline{Overview of Answer Set Optimization:}}
The paradigm of \emph{Answer Set Optimization} was originally
introduced by Brewka, Niemel\"{a}, and Truszczy\`{n}ski \cite{aso} and later 
refined by Brewka \cite{brewka-complex}. 

In ASO, a program is composed of two parts $\langle P_{gen}, P_{pref} \rangle$, where
$P_{gen}$ is an arbitrary logic program (the \emph{generator} program) and 
$P_{pref}$ is a collection of preference rules, used to define a preorder over
the answer sets of $P_{gen}$. The basic type of rules present in $P_{pref}$ are
of the form
\begin{equation}\label{form}
 C_1:p_1 > \dots > C_n: p_n \leftarrow Body 
\end{equation}
where $p_i$ are numerical weights while $C_j$ are propositional formulae. 
The complex types of preference rules in $P_{pref}$ is defined inductively
using rules of the form (\ref{form}) and the constructors 
{\em psum, inc, rinc, card, rcard, pareto,} and {\em lex}.

For each rule $r$ of the form (\ref{form}),
an answer set $S$ of the program $\langle P_{gen}, P_{pref} \rangle$
yields a penalty $pen(S,r)$ which is defined by 
(i) $pen(S,r) = p_j$ where $j = \min\{i \mid S \models C_i\}$ if 
$S$ satisfies $Body$ and at least one $C_i$, and 
(ii) $pen(S,r) = 0$ otherwise. This penalty is used in defining 
a preorder among answer sets of the program as follows.

Given two answer sets $S_1,S_2$ of an ASO program $P$, we have that $S_1$ is 
preferred to $S_2$ ($S_1 \geq S_2$) w.r.t.
a rule $r$ in $P$ of the form (\ref{form}) if 
\[ 
pen(S_1,r) \leq pen(S_2,r).
\]
More complex types of
preorder can be described by combining preference rules using a predefined set of 
constructors:
\begin{itemize}
\item $(psum\:\:e_1,\dots,e_k)$, where $S_1 \geq S_2$ iff 
\[\sum_{i=1}^k pen(S_1,e_i) \leq \sum_{i=1}^k pen(S_2,e_i)\]
\item $(rinc\:\:e_1,\dots,e_k)$, where $S_1 \geq S_2$ iff
\[ \exists 1 \leq i \leq k. (Pen^i(S_1) \supset Pen^i(S_2) \wedge
			\forall j < i. (Pen^j(S_1)=Pen^j(S_2))) \]
or 
\[ \forall 1 \leq i \leq k. (Pen^i(S_1) = Pen^i(S_2)) \]
where $Pen^i(S) = \{ j \:|\: pen(S,e_j) = i \}$.
\item $(rcard\:\:e_1,\dots,e_k)$, where $S_1 \geq S_2$ iff
\[\exists 1 \leq i \leq k. (|Pen^i(S_1)| > |Pen^i(S_2)| \wedge
			\forall j < i. (|Pen^j(S_1)|=|Pen^j(S_2)|)) \]
or 
\[ \forall 1 \leq i \leq k. (|Pen^i(S_1)| = |Pen^i(S_2)|) \]
\item $(lex\:\:e_1,\dots,e_k)$, where $S_1 \geq S_2$ iff
\[\exists 1 \leq i \leq k. ( S_1 >_i S_2 \wedge
			\forall j < i. (S_1 \geq_j S_2)) \]
or 
\[ \forall 1 \leq i \leq k. (S_1 \geq_i S_2)\]
where $\geq_i$ is the preorder associated to the expression $e_i$.
\item $(pareto\:\:e_1,\dots,e_k)$, where $S_1 \geq S_2$ iff
\[ \forall 1 \leq i \leq k. (S_1 \geq_i S_2) \]
\end{itemize}
where each $e_i$ is a preference rule in $P_{pref}$.

\paragraph{\underline{Encoding of our Preferences:}}
As for the case of LPOD, the encoding of our preference language in 
ASO is simple for the first two levels (basic desires and atomic preferences),
while it is more complex in the case of general preferences.

We will follow an encoding structure that is analogous to the one used in 
Section \ref{lpod}. In particular, we maintain the same definition of
$Core^{\psi}(T)$ and $rules^{\psi}(T)$. In this case, the generator program
$P_{gen}$ corresponds simply to the $\Pi(D,I,G)$ program that encodes the
planning problem. The preference rules employed in the various cases are
the following:
\begin{itemize}
\item if $\psi(T) \equiv occ(a)$ then
\[ e^{\psi}(T) \equiv occ(a,T) > \neg occ(a,T) \leftarrow .\]
\item if $\psi(T) \equiv f$ (where $f$ is a fluent literal) then
\[ e^{\psi}(T) \equiv holds(a,T) > \neg holds(a,T) \leftarrow .\]
\item if $\psi(T) \equiv \psi_1(T) \wedge \psi_2(T)$ then
\[ e^{\psi}(T) \equiv (Core^{\psi_1}(T)\wedge Core^{\psi_2}(T)) > \top \leftarrow.\]
(where $\top$ is a tautology).
\item if $\psi(T) \equiv \psi_1(T) \vee \psi_2(T)$ then
\[ e^{\psi}(T) \equiv (Core^{\psi_1}(T) \vee Core^{\psi_2}(T)) > \top \leftarrow .\]
\item if $\psi(T) \equiv \neg \psi_1(T)$ then
\[ e^{\psi}(T) \equiv \neg Core^{\psi_1}(T) > Core^{\psi_1}(T) \leftarrow.\]
\item if $\psi(T) \equiv \next(\psi_1(T))$ then
\[ e^{\psi}(T) \equiv Core^{\psi_1}(T+1) > \neg Core^{\psi_1}(T+1) \leftarrow .\]
\item if $\psi(t) \equiv \eventually(\psi_1(t))$ (with $1\leq t\leq n$, where $n$
	is the length of the desired plan) then
\[ e^{\psi}(t) \equiv 
	(Core^{\psi_1}(t) \vee Core^{\psi_1}(t+1) \vee \dots \vee Core^{\psi_1}(n)) > \top \leftarrow . \]
\item if $\psi(t) \equiv \always(\psi_1(t))$ then
\[ e^{\psi}(t) \equiv (Core^{\psi_1}(t) \wedge Core^{\psi_1}(t+1) \wedge \dots \wedge Core^{\psi_1}(n)) > \top \leftarrow . \]
\end{itemize}
With respect to the original definition of ASO, which allows for a ranked sequence
of preference programs,  atomic preferences of the type
$\psi_1 \lhd \psi_2$ can be encoded as $\langle \{e^{\psi_1}\}, \{e^{\psi_2}\} \rangle$.
In the extended ASO model proposed in \cite{brewka-complex}, the same effect can
be obtained by using the expression
\[ (pareto \:\: e^{\psi_1}, e^{\psi_2})\]

Only some of the general preferences can be directly encoded without 
relying on the use of numeric weights. 
\begin{itemize}
\item if $\psi \equiv \psi_1 \& \psi_2$, then we can introduce the expression
\[ e^{\psi} \equiv (pareto \:\: e^{\psi_1}, e^{\psi_2}) \]
\item if $\psi \equiv \psi_1 \lhd \psi_2$, then we can introduce the expression
\[ e^{\psi} \equiv (lex \:\: e^{\psi_1}, e^{\psi_2})\]
\end{itemize}
The other cases appear to require the use of weights, leading to an encoding
as complex as the one presented in Section \ref{implementation}.

For the cases listed above, we can assert the following result:
for a planning problem $\langle D,I,G\rangle$,  a preference  $\psi$, and 
two answer sets $S_1,S_2$ of $\Pi(D,I,G)$, it holds that 
\[ \begin{array}{lcl}
S_1 \geq_{\psi} S_2 & \textit{  iff  } &
	\pi(S_1) \preceq_{\psi} \pi(S_2)
   \end{array}
\]
where $\geq_{\psi}$ is the preorder derived from the expression $e^{\psi}$. 


\section{Conclusion and Future Work}
\label{sec5}
In this paper we presented a novel declarative language,
called ${\cal PP}$, for the specification of preferences
in the context of planning problems. The language nicely
integrates with  traditional action description languages
(e.g., $\cal B$) and it allows the elegant encoding of complex
preferences between trajectories. The language provides 
a \emph{declarative} framework for the encoding of preferences,
allowing users to focus on the high-level description of
preferences (more than their encodings---as in the approaches
based on utility functions). ${\cal PP}$ allows the expression
of complex preferences, including multi-dimensional preferences.
We also demonstrated
that ${\cal PP}$ preferences can be elegantly handled in
a logic programming framework based on answer set semantics.

The implementation of the language 
$\cal PP$ in the {\bf jsmodels} system is
almost complete, and this will offer us the opportunity
to validate our ideas on large test cases and to compare with 
related work such as that in \cite{eiter02a}. We would also like 
to develop a direct implementation of the language which can  
guarantee completeness. In other words, we would like to develop a system 
that can return \emph{all} possible 
preferred trajectories.

We also intend to explore the possibility of introducing temporal
operators at the level of general preferences. These seem to 
allow for very compact representation of various types of preferences;
for example, a shortest plan preference can be encoded simply as:
$$ 
\textbf{always}((occ(stop)\vee occ(noop)) \lhd (occ(a_1)\vee \dots \vee occ(a_k))) 
$$
if $a_1,\dots,a_k$ are the possible actions. We also intend to natively
include in the language preferences like $maxim$ used in 
Section \ref{planpref}; these preferences are already expressible in the
existing $\cal PP$ language, but at the expense of large and complex
preference formulae. Furthermore, we would like to develop a 
system that can assist users in defining the preferences 
given the planning problem. 

\subsection*{Acknowledgments} 
The authors wish to thank the anonymous referees for their
valuable comments. 
The authors were supported by 
the NSF grants CNS-0220590, CNS-0454066, and HRD-0420407.
The authors also wish to thank Hung Le for his implementation
of {\bf jsmodels}.


\begin{thebibliography}{}

\bibitem[\protect\citeauthoryear{Apt, Blair, and Walker}{Apt
  et~al\mbox{.}}{1988}]{apt88}
{\sc Apt, K.}, {\sc Blair, H.}, {\sc and} {\sc Walker, A.} 1988.
\newblock Towards a theory of declarative knowledge.
\newblock In {\em Foundations of Deductive Databases and Logic Programming}. 
Morgan Kaufmann, 89--148.

\bibitem[\protect\citeauthoryear{Bacchus and Kabanza}{Bacchus and
  Kabanza}{2000}]{bacchus00}
{\sc Bacchus, F.} {\sc and} {\sc Kabanza, F.} 2000.
\newblock Using temporal logics to express search control knowledge for
  planning.
\newblock {\em Artificial Intelligence\/}~{\em 116,\/}~1,2, 123--191.

\bibitem[\protect\citeauthoryear{Bistarelli, Codognet, Georget, and
  Rossi}{Bistarelli et~al\mbox{.}}{2000}]{bista1}
{\sc Bistarelli, S.}, {\sc Codognet, P.}, {\sc Georget, Y.}, {\sc and} {\sc
  Rossi, F.} 2000.
\newblock {Labeling and Partial Local Consistency for Soft Constraint
  Programming}.
\newblock In {\em Practical Aspects of Declarative Languages}. Springer Verlag,
  230--248.

\bibitem[\protect\citeauthoryear{Bistarelli, Montanari, and Rossi}{Bistarelli
  et~al\mbox{.}}{1997}]{bista-soft}
{\sc Bistarelli, S.}, {\sc Montanari, U.}, {\sc and} {\sc Rossi, F.} 1997.
\newblock {Semiring Based Constraint Solving and Optimization}.
\newblock {\em Journal of the ACM\/}~{\em 44,\/}~2, 201--236.

\bibitem[\protect\citeauthoryear{Blum and Furst}{Blum and
  Furst}{1997}]{blumfurst}
{\sc Blum, A.} {\sc and} {\sc Furst, M.} 1997.
\newblock {Fast Planning through Planning Graph Analysis}.
\newblock {\em Artificial Intelligence\/}~{\em 90}, 281--300.

\bibitem[\protect\citeauthoryear{Borning, Maher, Martindale, and
  Wilson}{Borning et~al\mbox{.}}{1989}]{borning-hierarchy}
{\sc Borning, A.}, {\sc Maher, M.}, {\sc Martindale, A.}, {\sc and} {\sc
  Wilson, M.} 1989.
\newblock {Constraint Hierarchies and Logic Programming}.
\newblock In {\em Proceedings of the International Conference on Logic
  Programming}. MIT Press, 149--164.

\bibitem[\protect\citeauthoryear{Boutilier, Brafman, Domshlak, Hoos, and
  Poole}{Boutilier et~al\mbox{.}}{2004}]{cpnet2}
{\sc Boutilier, C.}, {\sc Brafman, R.}, {\sc Domshlak, C.}, {\sc Hoos, H.},
  {\sc and} {\sc Poole, D.} 2004.
\newblock {Preference-based Constrained Optimization with CP-Nets}.
\newblock {\em Computational Intelligence\/}~{\em 20,\/}~2, 137--157.

\bibitem[\protect\citeauthoryear{Boutilier, Brafman, Hoos, and Poole}{Boutilier
  et~al\mbox{.}}{1999}]{cpnet}
{\sc Boutilier, C.}, {\sc Brafman, R.}, {\sc Hoos, H.}, {\sc and} {\sc Poole,
  D.} 1999.
\newblock {Reasoning with Conditional Ceteris Paribus Preference Statements}.
\newblock In {\em $15^{th}$ Annual Conference on Uncertainty in Artificial
  Intelligence}. Morgan Kaufmann, 71--80.

\bibitem[\protect\citeauthoryear{Brewka}{Brewka}{2004a}]{brewka-qualitative}
{\sc Brewka, G.} 2004a.
\newblock {A Rank Based Description Language for Qualitative Preferences}.
\newblock In {\em Proceedings of ECAI},  IOS Press, 303--307.

\bibitem[\protect\citeauthoryear{Brewka}{Brewka}{2004b}]{brewka-complex}
{\sc Brewka, G.} 2004b.
\newblock {Complex Preferences for Answer Set Optimization}.
\newblock In {\em Proceedings of KR}, AAAI Press, 213--223.

\bibitem[\protect\citeauthoryear{Brewka and Eiter}{Brewka and
  Eiter}{1999}]{brew99}
{\sc Brewka, G.} {\sc and} {\sc Eiter, T.} 1999.
\newblock Preferred answer sets for extended logic programs.
\newblock {\em Artificial Intelligence\/}~{\em 109}, 297--356.

\bibitem[\protect\citeauthoryear{Brewka, Niemel\"{a}, and Syrj\"{a}nen}{Brewka
  et~al\mbox{.}}{2002}]{lpod}
{\sc Brewka, G.}, {\sc Niemel\"{a}, I.}, {\sc and} {\sc Syrj\"{a}nen, T.} 2002.
\newblock {Implementing Ordered Disjunction using Answer Set Solvers for Normal
  Programs}.
\newblock In {\em Logics in Artificial Intelligence}. Springer Verlag,
  444--455.

\bibitem[\protect\citeauthoryear{Brewka, Niemel\"{a}, and
  Truszczy\`{n}ski}{Brewka et~al\mbox{.}}{2003}]{aso}
{\sc Brewka, G.}, {\sc Niemel\"{a}, I.}, {\sc and} {\sc Truszczy\`{n}ski, M.}
  2003.
\newblock {Answer Set Optimization}.
\newblock In {\em International Joint Conference on Artificial Intelligence}.
  Morgan Kaufmann, 867--872.

\bibitem[\protect\citeauthoryear{Buccafurri, Leone, and Rullo}{Buccafurri
  et~al\mbox{.}}{2000}]{dlv-constraints}
{\sc Buccafurri, F.}, {\sc Leone, N.}, {\sc and} {\sc Rullo, P.} 2000.
\newblock {Enhancing Disjunctive Datalog by Constraints}.
\newblock {\em IEEE Transactions on Knowledge and Data Engineering\/}~{\em
  12,\/}~5, 845--860.

\bibitem[\protect\citeauthoryear{Cimatti and Roveri}{Cimatti and
  Roveri}{2000}]{cimatti}
{\sc Cimatti, A.} {\sc and} {\sc Roveri, M.} 2000.
\newblock {Conformant Planning via Symbolic Model Checking}.
\newblock {\em Journal of Artificial Intelligence Research\/}~{\em 13},
  305--338.

\bibitem[\protect\citeauthoryear{Cui and Swift}{Cui and Swift}{2002}]{cui}
{\sc Cui, B.} {\sc and} {\sc Swift, T.} 2002.
\newblock {Preference Logic Grammars: Fixed Point Semantics and Application to
  Data Standardization}.
\newblock {\em Artificial Intelligence\/}~{\em 138,\/}~1--2, 117--147.


\bibitem[\protect\citeauthoryear{Dal Lago, Pistore, and Traverso}
{Dal Lago, Pistore, and Traverso}{2002}]{lago}
{\sc Dal Lago, U.} {\sc and} 
{\sc Pistore, M.} {\sc and} 
{\sc Traverso,  P.} 2002. 
\newblock {Planning with a Language for Extended Goals}. 
\newblock In {\em Proceedings of AAAI/IAAI}. AAAI Press, 447--454.

\bibitem[\protect\citeauthoryear{Delgrande, Schaub, and Tompits}{Delgrande
  et~al\mbox{.}}{2003}]{delschtomp03}
{\sc Delgrande, J.}, {\sc Schaub, T.}, {\sc and} {\sc Tompits, H.} 2003.
\newblock A framework for compiling preferences in logic programs.
\newblock {\em Theory and Practice of Logic Programming\/}~{\em 3,\/}~2 (Mar.),  
  129--187.

\bibitem[\protect\citeauthoryear{Delgrande, Schaub, and Tompits}{Delgrande
  et~al\mbox{.}}{2004}]{delgrande-plan}
{\sc Delgrande, J.}, {\sc Schaub, T.}, {\sc and} {\sc Tompits, H.} 2004.
\newblock {Domain-specific Preferences for Causal Reasoning and Planning}.
\newblock In {\em Principles of Knowledge Representation and Reasoning}. AAAI
  Press, 673--682.

\bibitem[\protect\citeauthoryear{Dimopoulos, Nebel, and Koehler}{Dimopoulos
  et~al\mbox{.}}{1997}]{dimo97}
{\sc Dimopoulos, Y.}, {\sc Nebel, B.}, {\sc and} {\sc Koehler, J.} 1997.
\newblock Encoding planning problems in non-monotonic logic programs.
\newblock In {\em Proceedings of European Conference on Planning}. 
Springer Verlag, 169--181.

\bibitem[\protect\citeauthoryear{Domshlak and Brafman}{Domshlak and
  Brafman}{2002}]{cpnet1}
{\sc Domshlak, C.} {\sc and} {\sc Brafman, R.} 2002.
\newblock {CP-Nets: Reasoning and Consistency Testing}.
\newblock In {\em $8^{th}$ International Conference on Principles of Knowledge
  Representation and Reasoning}. Morgan Kaufmann, 121--132.

\bibitem[\protect\citeauthoryear{Eiter, Faber, Leone, Pfeifer, and
  Polleres}{Eiter et~al\mbox{.}}{2003}]{eiter02a}
{\sc Eiter, T.}, {\sc Faber, W.}, {\sc Leone, N.}, {\sc Pfeifer, G.}, {\sc and}
  {\sc Polleres, A.} 2002.
\newblock {Answer Set Planning under Action Cost}.
\newblock {\em Journal of Artificial Intelligence Research}, {\em 19}, 25--71.

\bibitem[\protect\citeauthoryear{Fargier and Lang}{Fargier and
  Lang}{1993}]{fargier}
{\sc Fargier, H.} {\sc and} {\sc Lang, J.} 1993.
\newblock {Uncertainty in Constraint Satisfaction Problems: a Probabilistic
  Approach}.
\newblock In {\em European Conference on Symbolic and Qualitative Approaches to
  Reasoning and Uncertainty}. Springer Verlag, 97--104.

\bibitem[\protect\citeauthoryear{Gelfond and Lifschitz}{Gelfond and
  Lifschitz}{1998}]{gl98}
{\sc Gelfond, M.} {\sc and} {\sc Lifschitz, V.} 1998.
\newblock Action languages.
\newblock {\em Electron. Trans. Artif. Intell.\/}~{\em 2}: 193-210.

\bibitem[\protect\citeauthoryear{Gelfond, Przymusinska, and
  Przymusinski}{Gelfond et~al\mbox{.}}{1990}]{gel90a}
{\sc Gelfond, M.}, {\sc Przymusinska, H.}, {\sc and} {\sc Przymusinski, T.}
  1990.
\newblock {On the relationship between CWA, Minimal Model, and Minimal Herbrand
  Model semantics}.
\newblock {\em International Journal of Intelligent Systems\/}~{\em 5,\/}~5,
  549--565.

\bibitem[\protect\citeauthoryear{Haddawy and Hanks}{Haddawy and
  Hanks}{1993}]{haddawy93a}
{\sc Haddawy, P.} {\sc and} {\sc Hanks, S.} 1993.
\newblock {Utility Model for Goal-Directed Decision Theoretic Planners}.
\newblock Tech. Rep., University of Washington.

\bibitem[\protect\citeauthoryear{Junker}{Junker}{2001}]{junker-config}
{\sc Junker, U.} 2001.
\newblock {Preference Programming for Configuration}.
\newblock In {\em Proceedings of the IJCAI Workshop on Configuration}.
\newblock \url{www.soberit.hut.fi/pdmg/IJCAI2001ConfWS/S}.

\bibitem[\protect\citeauthoryear{Kautz and Walser}{Kautz and
  Walser}{1999}]{kautz1}
{\sc Kautz, H.} {\sc and} {\sc Walser, J.} 1999.
\newblock {State-space Planning by Integer Optimization}.
\newblock In {\em Proceedings of AAAI}. AAAI Press, 526--533.

\bibitem[\protect\citeauthoryear{Larrosa}{Larrosa}{2002}]{larrosa}
{\sc Larrosa, J.} 2002.
\newblock {On Arc and Node Consistency in Weighted CSP}.
\newblock In {\em Proceedings of AAAI}. AAAI Press, 48--53. 

\bibitem[\protect\citeauthoryear{Larrosa and Schiex}{Larrosa and
  Schiex}{2003}]{schiex2}
{\sc Larrosa, J.} {\sc and} {\sc Schiex, T.} 2003.
\newblock {In the Quest of the Best Form of Local Consistency for Weighted
  CSP}.
\newblock In {\em Proceedings of IJCAI}. Morgan Kaufmann, 239--244.

\bibitem[\protect\citeauthoryear{Le and Pontelli}{Le and
  Pontelli}{2003}]{jsmodels}
{\sc Le, H.~V.} {\sc and} {\sc Pontelli, E.} 2003.
\newblock {A Java Based Solver for Answer Set Programming}.
\newblock \url{www.cs.nmsu.edu/lldap/jsmodels}.

\bibitem[\protect\citeauthoryear{Leone, Pfeifer, Faber, Eiter, Gottlob, 
  Perri, and  Scarcello}{Leone et~al\mbox{.}}{2005}]{eiter98a}
{\sc Leone, N.},  {\sc Pfeifer, G.},  {\sc Faber, W.},
{\sc Eiter, T.},  {\sc Gottlob, G.}, {\sc Perri, S.}
 {\sc  and} {\sc Scarcello, F.} 2005.
\newblock {The DLV System for Knowledge Representation and Reasoning}.
\newblock In {\em ACM Transaction on Computational Logic}. To Appear.

\bibitem[\protect\citeauthoryear{Lierler and Maratea}{Lierler and
  Maratea}{2004}]{cmodels2}
{\sc Lierler, Y.} {\sc and} {\sc Maratea, M.} 2004.
\newblock {Cmodels-2: SAT-based Answer Set Solver Enhanced to Non-tight
  Programs}.
\newblock In {\em Proceedings of the 7th International Conference on Logic
  Programming and Non-Monotonic Reasoning Conference (LPNMR'04)}. 
Springer Verlag, 346--350.

\bibitem[\protect\citeauthoryear{Lifschitz}{Lifschitz}{2002}]{lif02a}
{\sc Lifschitz, V.} 2002.
\newblock {Answer set programming and plan generation}.
\newblock {\em Artificial Intelligence\/}~{\em 138,\/}~1--2, 39--54.

\bibitem[\protect\citeauthoryear{Lifschitz and Turner}{Lifschitz and
  Turner}{1994}]{lif94a}
{\sc Lifschitz, V.} {\sc and} {\sc Turner, H.} 1994.
\newblock Splitting a logic program.
\newblock In {\em Proceedings~of the Eleventh International Conf.~on Logic
  Programming}. MIT Press, 23--38.

\bibitem[\protect\citeauthoryear{Lin}{Lin}{1998}]{lin98}
{\sc Lin, F.} 1998.
\newblock {On Measuring Plan Quality (A Preliminary Report)}.
\newblock In {\em Proceedings of the Sixth International Conferences on
  Principles of Knowledge Representation and Reasoning (KR'98)}. 224--233.

\bibitem[\protect\citeauthoryear{Lin and Zhao}{Lin and Zhao}{2002}]{assat}
{\sc Lin, F.} {\sc and} {\sc Zhao, Y.} 2002.
\newblock {ASSAT: Computing Answer Sets of A Logic Program By SAT Solvers}.
\newblock In {\em Proceedings of AAAI}. AAAI Press, 112--117.

\bibitem[\protect\citeauthoryear{Long, Fox, Smith, McDermott, Bacchus, and
  Geffner}{Long et~al\mbox{.}}{}]{aips02} 
{\sc Long, D.}, {\sc Fox, M.}, {\sc Smith, D.}, {\sc McDermott, D.}, {\sc
  Bacchus, F.}, {\sc and} {\sc Geffner, H.} 2002.
\newblock {International Planning Competition}.
\url{http://planning.cis.strath.ac.uk/competition/}

\bibitem[\protect\citeauthoryear{Moulin}{Moulin}{1988}]{moulin88}
{\sc Moulin, H.} 1988.
\newblock {\em {Axioms for Cooperative Decision Making}}.
\newblock Cambridge University Press.

\bibitem[\protect\citeauthoryear{Myers}{Myers}{1996}]{myers1}
{\sc Myers, K.} 1996.
\newblock {Strategic Advice for Hierarchical Planners}.
\newblock In {\em Principles of Knowledge Representation and Reasoning}. 
  Morgan
  Kaufmann, 112--123.

\bibitem[\protect\citeauthoryear{Myers and Lee}{Myers and Lee}{1999}]{myers99a}
{\sc Myers, K.} {\sc and} {\sc Lee, T.} 1999.
\newblock {Generating Qualitatively Different Plans through Metatheoretic
  Biases}.
\newblock In {\em Proceedings of AAAI}. AAAI Press, 570--576.

\bibitem[\protect\citeauthoryear{Niemel{\"{a}}}{Niemel{\"{a}}}{1999}]{smodels-%
constraint}
{\sc Niemel{\"{a}}, I.} 1999.
\newblock Logic programming with stable model semantics as a constraint
  programming paradigm.
\newblock {\em Annals of Mathematics and Artificial Intelligence\/}~{\em
  25,\/}~3,4, 241--273.

\bibitem[\protect\citeauthoryear{Przymusinski}{Przymusinski}{1988}]{prz88b}
{\sc Przymusinski, T.} 1988.
\newblock On the declarative semantics of deductive databases and logic
  programs.
\newblock In {\em Foundations of Deductive Databases and Logic Programming}. 
	Morgan Kaufmann, 193--216.

\bibitem[\protect\citeauthoryear{Putterman}{Putterman}{1994}]{puterman94}
{\sc Putterman, M.} 1994.
\newblock {\em Markov Decision Processes -- Discrete Stochastic Dynamic
  Programming}.
\newblock John Wiley \& Sons, Inc., New York, NY.

\bibitem[\protect\citeauthoryear{Reiter}{Reiter}{2001}]{Rei01}
{\sc Reiter, R.} 2001.
\newblock {\em {KNOWLEDGE IN ACTION}: Logical Foundations for Describing and
  Implementing Dynamical Systems}.
\newblock MIT Press.

\bibitem[\protect\citeauthoryear{Schaub and Wang}{Schaub and
  Wang}{2001}]{schaub}
{\sc Schaub, T.} {\sc and} {\sc Wang, K.} 2001.
\newblock {A Comparative Study of Logic Programs with Preferences}.
\newblock In {\em IJCAI}. Morgan Kaufman, 597--602.

\bibitem[\protect\citeauthoryear{Schiex and Cooper}{Schiex and
  Cooper}{2002}]{schiex-survey}
{\sc Schiex, T.} {\sc and} {\sc Cooper, M.} 2002.
\newblock {Constraints and Preferences: The Interplay of Preferences and
  Algorithms}.
\newblock In {\em Proceedings of the AAAI Workshop on Preferences in AI and
  CP}.

\bibitem[\protect\citeauthoryear{Schiex, Fargier, and Verfaillie}{Schiex
  et~al\mbox{.}}{1995}]{schiex1}
{\sc Schiex, T.}, {\sc Fargier, H.}, {\sc and} {\sc Verfaillie, G.} 1995.
\newblock {Valued Constraint Satisfaction Problems: Hard and Easy Problems}.
\newblock In {\em Proceedings of IJCAI}. Morgan Kaufmann, 631--637.

\bibitem[\protect\citeauthoryear{Simons, Niemel\"{a}, and Soininen}{Simons
  et~al\mbox{.}}{2002}]{smodels-ai}
{\sc Simons, P.}, {\sc Niemel\"{a}, I.}, {\sc and} {\sc Soininen, T.} 2002.
\newblock {Extending and Implementing the Stable Model Semantics}.
\newblock {\em Artificial Intelligence\/}~{\em 138,\/}~1--2, 181--234.

\bibitem[\protect\citeauthoryear{Son, Baral, Nam, and McIlraith}{Son
  et~al\mbox{.}}{2005}]{sbm02a}
{\sc Son, T.}, {\sc Baral, C.}, {\sc Nam, T.}, {\sc and} {\sc McIlraith, S.}
  2005.
\newblock {Domain-Dependent Knowledge in Answer Set Planning}.
\newblock {\em ACM Transaction on Computational Logic}. To Appear.

\bibitem[\protect\citeauthoryear{Son and Pontelli}{Son and
  Pontelli}{2004a}]{son-prio}
{\sc Son, T.} {\sc and} {\sc Pontelli, E.} 2004a.
\newblock {Reasoning about Actions and Planning with Preferences using
  Prioritized Default Theory}.
\newblock {\em Computational Intelligence\/}~{\em 20,\/}~2, 358--404.

\bibitem[\protect\citeauthoryear{Son and Pontelli}{Son and
  Pontelli}{2004b}]{son-prio-plan}
{\sc Son, T.} {\sc and} {\sc Pontelli, E.} 2004b.
\newblock {Planning with Preferences using Logic Programming}.
\newblock {\em Logic programming and Non-monotonic Reasoning\/}, Springer Verlag, 247--260.

\bibitem[\protect\citeauthoryear{Subrahmanian and Zaniolo}{Subrahmanian and
  Zaniolo}{1995}]{sub95}
{\sc Subrahmanian, V.} {\sc and} {\sc Zaniolo, C.} 1995.
\newblock Relating stable models and ai planning domains.
\newblock In {\em Proceedings of the International Conference on Logic
  Programming}. MIT Press, 233--247.

\end{thebibliography}
\end{document}